\documentclass[sigconf]{acmart}

\usepackage{array}
\usepackage{makecell}
\usepackage{multirow}
\usepackage{colortbl}
\usepackage{enumitem}
\usepackage{xurl}
\usepackage{cleveref}

\copyrightyear{2026}
\acmYear{2026}
\setcopyright{cc}
\setcctype{by-nc-nd}
\acmConference[CVMP '26]{European Conference on Visual Media Production}{December 08--09, 2026}{London, United Kingdom}
\acmBooktitle{European Conference on Visual Media Production (CVMP '26), December 08--09, 2026, London, United Kingdom}
\acmDOI{10.1145/3828644.3857002}
\acmISBN{979-8-4007-2823-5/2026/12}

\begin{document}

\title{SERO: Semantic Prediction for Single-Step Object Removal}


\author{Amreen Hasheem Siraj}
\authornote{These authors contributed equally to this work.}
\affiliation{%
  \institution{Rheinland-Pfälzische Technische Universität}
  \city{Kaiserslautern}
  \country{Germany}
}
\email{amreen.hasheem@gmail.com}
\orcid{0009-0002-5413-1607}

\author{Alireza Javanmardi}
\authornotemark[1]
\affiliation{%
  \institution{Deutsches Forschungszentrum für Künstliche Intelligenz}
  \city{Kaiserslautern}
  \country{Germany}
}
\email{alireza.javanmardi@dfki.de}
\orcid{0009-0008-4926-1566}

\author{Pragati Jaiswal}
\authornotemark[1]
\affiliation{%
  \institution{Rheinland-Pfälzische Technische Universität}
  \city{Kaiserslautern}
  \country{Germany}
}
\email{pragati.jaiswal@dfki.de}
\orcid{0000-0003-0422-0695}

\author{Didier Stricker}
\affiliation{%
\institution{Deutsches Forschungszentrum für Künstliche Intelligenz}
\city{Kaiserslautern}
\country{Germany}
}
\affiliation{%
  \institution{Rheinland-Pfälzische Technische Universität}
  \city{Kaiserslautern}
  \country{Germany}
}
\email{didier.stricker@dfki.de}
\orcid{0009-0004-8794-6858}

\author{Alain Pagani}
\affiliation{%
  \institution{Deutsches Forschungszentrum für Künstliche Intelligenz}
  \city{Kaiserslautern}
  \country{Germany}
}
\email{alain.pagani@dfki.de}
\orcid{0000-0002-5136-0837}


\begin{abstract}
Object removal and semantic inpainting of missing regions remain inherently ill-posed, as multiple plausible completions may exist for a given scene. Recent diffusion-based methods have demonstrated strong potential for high-quality object removal; however, they still face two key limitations. First, although they can synthesise sharp and high-resolution images, they often struggle to complete removed regions in a semantically consistent and contextually meaningful manner, where the generated content should agree with the surrounding scene layout and object relationships rather than merely appear visually realistic. Second, their iterative denoising process makes them computationally expensive, limiting their practicality for downstream applications. We propose SERO, a semantic prediction framework for single-step object removal. Our model predicts a hidden-region semantic prior in embedding space, which guides a latent bridge-matching generator toward plausible scene completion by providing both where to edit and what structure is likely to exist behind the removed object. This enables visually realistic and contextually coherent completions in one forward pass. Experiments on the RORD and OBER datasets show that SERO achieves competitive or state-of-the-art performance while requiring only a single generation step.\\
\noindent\textbf{Project page:}
\href{https://dfki-av.github.io/SERO/}{https://dfki-av.github.io/SERO/}
\end{abstract}





\begin{teaserfigure}
  \centering
  \includegraphics[width=\textwidth]{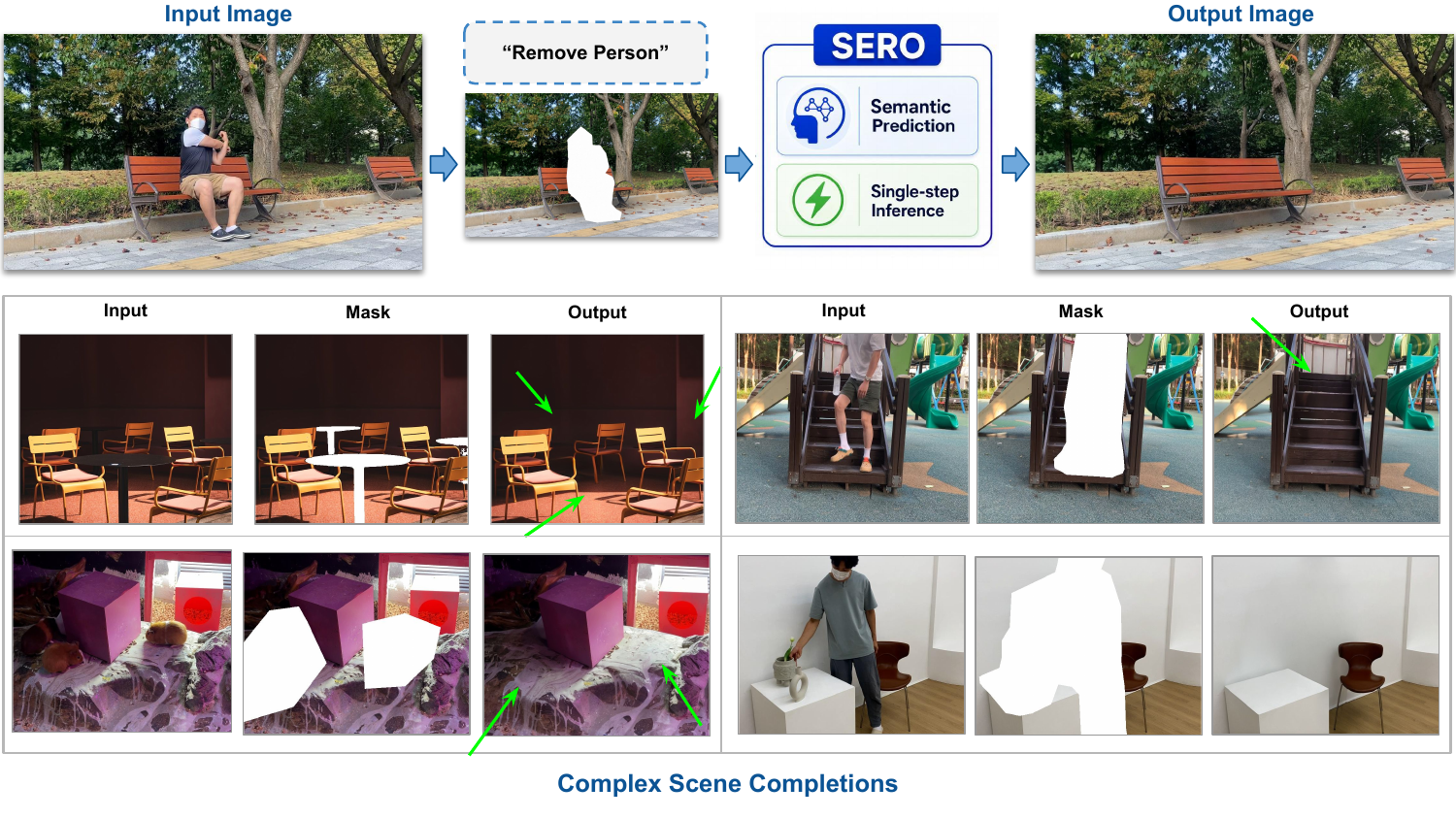}
  \Description{
    An overview and visual examples of the SERO single-step object-removal
    method.
  }
  \caption{
    SERO combines semantic prediction with efficient bridge matching to
    remove objects and reconstruct plausible scene content in a single step,
    across both real-world object removal and complex scene-completion
    scenarios.
    }
  \label{fig:teaser}
\end{teaserfigure}

\maketitle


\section{Introduction}
\label{sec:intro}
Generative models have transformed the field of image editing by enabling a wide range of content manipulation tasks, including image synthesis \cite{rombach2022high, peebles2023scalable}, object insertion \cite{yu2025omnipaint}, and object removal \cite{zhao2026objectclear}. Among these tasks, object removal is particularly important because of its broad practical applications, ranging from everyday photo editing on smartphones to specialized scenarios such as cloud removal in satellite imagery \cite{wang2024idf, liu2025effective}. The goal of object removal is not only to erase an unwanted object, but also to plausibly reconstruct the missing region such that the edited image remains visually coherent and indistinguishable from an unedited scene.

Early generative approaches, particularly GAN-based methods, advanced image inpainting and editing, but often struggled to synthesize high-resolution, photorealistic, and semantically consistent content \cite{yu2018generative, Nazeri_2019_ICCV}. More recently, diffusion models have shown remarkable image-generation capability and have become strong backbones for image inpainting \cite{rombach2022high, zhuang2024powerpaint}. Despite this progress, diffusion-based object removal methods still face critical challenges in complex scenes. Existing methods can perform well when the background is simple and the removed object has limited interaction with its surroundings. However, in scenes with multiple objects, structured layouts, and rich contextual dependencies, meaningful completion becomes substantially more difficult \cite{suvorov2022resolution}. In such cases, filling the removed region requires more than low-level texture synthesis; it requires understanding the scene structure, object relationships, and plausible semantic content that should replace the removed area. From this perspective, effective object removal should combine high-quality generation with semantic prediction about the missing content.

In this context, predictive representation learning approaches \cite{assran2023self} offer a promising direction for obtaining rich semantic priors in computer vision. Rather than reconstructing pixels directly, such methods predict the representation of a missing or masked region from the visible context, encouraging the model to capture higher-level scene structure and contextual dependencies. This formulation is particularly useful for object removal, where the goal is not only to fill missing pixels but also to infer plausible content that is consistent with surrounding objects, layout, and occlusion patterns. Motivated by these advantages, we explore how predictive semantic representations can guide object removal models toward more meaningful and contextually consistent inpainting. 

To overcome these limitations, we propose \textsc{SERO} (Semantic Prediction for Single-Step Object Removal), a novel framework that combines semantic prediction with efficient generative modelling for object removal. Our method integrates contextual prediction with a latent bridge-matching generative backbone, enabling the model to reason about the content that should plausibly replace a removed object rather than merely filling the masked region with low-level texture. By predicting semantic guidance in the embedding space, our method encourages the removal of both the target object and its associated visual effects, including shadows, reflections, and occlusion artifacts. This guidance is then used to steer latent bridge matching toward photorealistic and contextually consistent inpainting, achieving high-quality object removal in as few as one generation step. In summary, our main contributions are:

\begin{itemize}
\item We introduce \textsc{SERO}, a one-step object removal framework that produces photorealistic edits while removing not only the target object but also its associated visual effects, such as shadows and reflections.

\item We propose a predictive semantic-prior bridge matching architecture that integrates contextual semantic prediction with an efficient generative backbone, enabling more meaningful and context-aware completion of missing regions in complex and occluded scenes.

\item We evaluate \textsc{SERO} on the RORD and OBER benchmarks, which include challenging object-removal cases with object-associated effects, and demonstrate consistent improvements in both quantitative metrics and qualitative comparisons.
\end{itemize}

\section{Related Work}
\subsection{Image Inpainting and Generative Completion}
Image inpainting synthesises plausible content in missing or user-specified regions while preserving consistency with the surrounding context. Early learning-based methods used adversarial training and encoder-decoder architectures to hallucinate missing pixels from contextual cues~\cite{goodfellow2014generative,pathak2016context,liu2020rethinking}. Later works introduced structural priors, multi-scale context aggregation, and appearance-flow reasoning to better preserve edges, layouts, and long-range dependencies~\cite{ren2019structureflow,zeng2019learning}. However, GAN-based methods often struggle with large missing regions, complex textures, and diverse plausible completions.

Diffusion and flow-based generative models have become strong priors for image synthesis and restoration~\cite{ho2020denoising, lipman2022flow}. Denoising diffusion and score-based models~\cite{ho2020denoising,song2020score} enable iterative refinement, while latent diffusion supports efficient high-resolution generation~\cite{rombach2022high,podell2024sdxl}. Recently, flow-matching and rectified-flow models, such as Stable Diffusion 3~\cite{esser2024scaling} and FLUX \cite{labs2025flux}, have shown strong image generation quality by learning continuous transport paths between noise and data distributions. These priors have been adapted to editing and inpainting through stochastic editing, resampling-based reconstruction, text-guided blending, and task-conditioned prompting~\cite{meng2021sdedit,lugmayr2022repaint,avrahami2022blended,zhuang2024powerpaint}. Recent methods further improve controllability by separating masked and unmasked branches or reducing hallucinations~\cite{ju2024brushnet,wang2025enhanced}. Still, generic inpainting fills the given hole with plausible content, whereas object removal requires the image to appear as if the target object, including its scene influence, never existed.

\subsection{Object Removal}
Object removal is a specialised inpainting task that eliminates a target object while reconstructing the underlying scene. Recent methods often rely on diffusion priors. Some perform zero-shot or training-free removal with attention guidance, latent decomposition, or test-time optimization~\cite{chen2024freecompose,jia2025designedit,sun2025attentive,ekin2024clipaway}. While avoiding task-specific training, they can be sensitive to masks, prompts, and optimisation settings. Other methods train dedicated removal models using synthetic or curated paired data. SmartEraser~\cite{jiang2025smarteraser} and EraseDiffusion~\cite{liu2025erasediffusion} build scalable data with segmentation, matting, or diffusion-based pipelines, while RORem~\cite{li2025rorem} incorporates human feedback for robustness and annotation quality. More recent methods such as OmniEraser~\cite{wei2025omnieraser}, ObjectClear~\cite{zhao2026objectclear}, and GeoRemover~\cite{zhu2025georemover} explicitly address object-associated effects, including shadows, reflections, and other causal artifacts.

Despite these advancements, object-and-effect removal remains difficult in complex scenes where targets occlude background content or interact with surrounding objects. The model must remove associated effects and complete the scene with plausible structure. Moreover, iterative or multi-stage diffusion pipelines are often slow, limiting real-world applicability. We address these challenges with a semantic-prior-guided Latent Bridge Matching framework for coherent single-step object removal.
\section{Method}
\label{sec:method}

In our SERO framework, we remove a target object and its associated visual effects from a source image using a binary mask that specifies the region to be erased. Our approach first predicts a semantic representation of the hidden scene from the visible context and transforms it into a dense spatial prior that guides latent bridge matching for plausible background reconstruction. We first formulate the object removal problem in Sec.~\ref{sec:problem}, followed by the Latent Bridge Matching backbone in Sec.~\ref{sec:lbm}. We then introduce the predictive semantic prior and its integration into the generative model in Sec.~\ref{jepa}, before describing the training objective and inference procedure in Sec.~\ref{sec:train}.

\begin{figure*}[t]
  \centering
  \includegraphics[width=\linewidth]{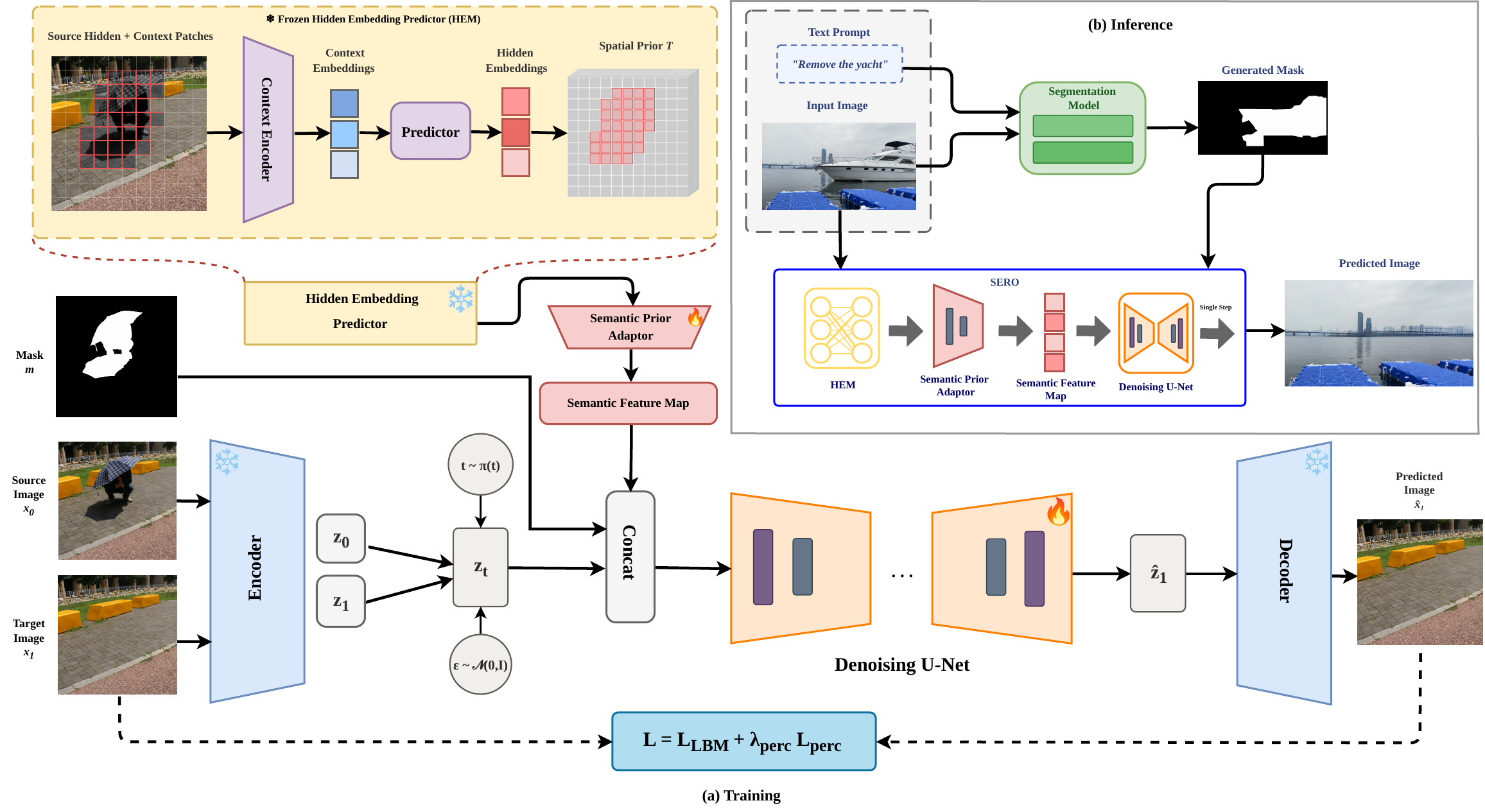}
  \caption{%
    Overview of the proposed pipeline.
    \textbf{(a) Training:} the frozen context encoder and predictor produce hidden-region token embeddings from visible background patches, giving us a spatial prior; a trainable decoder adapter maps this to a dense latent prior that is concatenated with the bridge latent and mask before being fed to the Denoising U-Net.
    \textbf{(b) Inference:} given a text prompt and input image, a segmentation model generates the removal mask, after which the semantic-prior-conditioned SERO model predicts the clean output in a single forward pass.}
  \Description{The SERO training and inference pipelines. During training, a frozen context encoder and predictor estimate hidden-region embeddings, an adaptor maps them to a spatial prior, and a U-Net learns object removal. During inference, a prompt-derived mask and the semantic prior guide a single-step clean-image prediction.}
  \label{fig:arch}
\end{figure*}

\subsection{Problem Formulation}
\label{sec:problem}

Let $x_0 \in \mathbb{R}^{H \times W \times 3}$ denote a source image containing an unwanted object or visual artifact, and let $m \in \{0,1\}^{H \times W}$ be the corresponding binary removal mask. We define $m_{ij}=0$ for pixels inside the region to be removed and $m_{ij}=1$ for pixels belonging to the valid background. The target image $x_1 \in \mathbb{R}^{H \times W \times 3}$ represents the clean ground-truth image in which the masked content has been removed. At inference time, given only $(x_0,m)$, the objective is to synthesise an edited image $\hat{x}_1$ whose masked region is plausibly reconstructed while the unmasked content remains faithful to the source image.

\subsection{Latent Bridge Matching Backbone}
\label{sec:lbm}

We build upon Latent Bridge Matching (LBM)~\cite{chadebec2025lbm}, which formulates image-to-image translation as learning a stochastic transport process between source and target image distributions. Both source and target images are first encoded using a pretrained SDXL variational autoencoder \cite{podell2024sdxl}:

\begin{equation}
z_0=\mathcal{E}(x_0),
\qquad
z_1=\mathcal{E}(x_1),
\end{equation}

where $z_0,z_1 \in \mathbb{R}^{4 \times h \times w}$ are latent representations and $(h,w)=(H/8,W/8)$.

Following LBM, an intermediate bridge latent is constructed as

\begin{equation}
z_t=(1-t)z_0+t z_1+\sigma \sqrt{t(1-t)}\,\varepsilon,
\qquad
\varepsilon \sim \mathcal{N}(0,I),
\end{equation}
where $t \in [0,1]$ is the bridge timestep and $\sigma$ controls the bridge noise level.

The denoising UNet is trained to predict the velocity field that transports the bridge latent toward the target latent. For object removal, the model is conditioned on the bridge latent together with a downsampled removal mask, which is resized to match the latent resolution and concatenated with the latent representation.

Although this conditioning effectively specifies the spatial region to be edited, it provides no explicit guidance regarding the content that should replace the removed object. As a result, the model must infer the missing scene structure solely from the surrounding visual context during generation. To overcome this limitation, we augment the conditioning representation with a predictive semantic prior, providing high-level semantic guidance for plausible content reconstruction.


\subsection{Predictive Semantic Prior}
\label{jepa}
Our key idea is to estimate a high-level semantic representation of the hidden scene before generative reconstruction. Following the predictive representation learning paradigm of I-JEPA~\cite{assran2023self}, this formulation is particularly well-suited to object removal, as it predicts the missing region from visible context in representation space rather than directly reconstructing pixels.
\subsubsection{Hidden Embedding Predictor}
Given the input image $x_0$, we divide it into a regular patch grid of size $G \times G$. The removal mask is mapped to this grid, producing hidden and visible patch sets:
\begin{align}
  I_{\text{hid}} &= \{\,i : \hat{m}_i = 0\,\}, \notag \\
  I_{\text{ctx}} &= \{\,i : \hat{m}_i = 1\,\}.
  \label{eq:patch-split}
\end{align}

Let $x_J$ be the resized image fed to the frozen Context Encoder, which produces visible-patch embeddings
\begin{equation}
  \mathbf{e}_{\text{ctx}} = \mathcal{E}_{J}(x_J)\bigl[I_{\text{ctx}}\bigr]
  \;\in\; \mathbb{R}^{|I_{\text{ctx}}| \times d},
  \label{eq:ctx-embed}
\end{equation}
The Predictor model  $\mathcal{P}_{J}$ then estimates the hidden-region representations from the context embedding, hidden and visible patch sets:
\begin{equation}
  \hat{\mathbf{h}} = \mathcal{P}_{J}\!\left(\mathbf{e}_{\text{ctx}},\, I_{\text{ctx}},\, I_{\text{hid}}\right)
  \;\in\; \mathbb{R}^{|I_{\text{hid}}| \times d}.
  \label{eq:jepa-pred}
\end{equation}
These predicted embeddings form a semantic prior over the content likely to occupy the removed region.

\subsubsection{Spatial Prior Construction}

To align the semantic prediction with the generative backbone, we scatter the hidden-region embeddings into a complete token grid:
\begin{equation}
\mathbf T \in \mathbb R^{G \times G \times d},
\end{equation}
where predicted tokens are placed at hidden locations, and all context locations are set to zero:
\begin{equation}
\mathbf{T}[I_{\text{hid}}] = \hat{\mathbf{h}}, \qquad
\mathbf{T}[I_{\text{ctx}}] = \mathbf{0}.
\end{equation}
Context positions are set to zero to avoid redundancy with the bridge latent and to focus the prior on the missing region. 

This representation preserves spatial correspondence with the original image while encoding the predicted structure of the missing region.

\subsubsection{Semantic Prior Adaptor}
The spatial prior $\mathbf{T}$ lies in the ViT embedding space and is not directly compatible with the U-Net input resolution. We therefore introduce a lightweight Semantic Prior Adaptor, denoted by $\mathcal{D}_{\mathrm{adapt}}$, which maps $\mathbf{T}$ into the latent conditioning space:
\begin{equation}
\mathbf{f}_J = \mathcal{D}_{\mathrm{adapt}}(\mathbf{T}),
\qquad
\mathbf{f}_J \in \mathbb{R}^{C \times h \times w}.
\label{eq:semantic-prior-adaptor}
\end{equation}
The adapter uses a 1×1 convolution to reduce channels from $d=1280$ to 128, followed by two bilinear upsampling stages with 3×3 convolutions, mapping to $C=4$ conditioning channels at the latent resolution $(h,w)$. Unlike an image decoder, its goal is not pixel synthesis but translation of semantic predictions into a conditioning tensor.

\subsubsection{Conditioning the Generative Model}
The semantic feature map 
$\mathbf{f}_J = \mathcal{D}_{\mathrm{adapt}}(\mathbf{T}) 
\in \mathbb{R}^{C \times h \times w}$ 
is concatenated with the mask conditioning to form the full conditioning signal:
\begin{equation}
  \mathbf{c} = \bigl[\;\hat{m}\;;\;\mathbf{f}_{J}\,\bigr]
  \;\in\; \mathbb{R}^{(1+C) \times h \times w}.
  \label{eq:full-cond}
\end{equation}
This condition provides both location information, through the mask, and structural guidance, through the semantic prior. The first convolutional layer of the UNet is expanded to accept the additional conditions, while the remaining parameters are initialized from the pretrained SDXL checkpoint \cite{podell2024sdxl}. 

\begin{table*}[!t]
\centering
\scriptsize
\setlength{\tabcolsep}{4.8pt}
\caption{\textbf{Quantitative results on RORD-Val and OBER-Test.}
We report FID, localFID, fMSE, PSNR$_{\text{FG}}$, LPIPS, and CLIP distance across both benchmark datasets.
\textbf{Bold} denotes best and \underline{underline} denotes second best per column.}
\renewcommand{\arraystretch}{1.0}
\resizebox{\textwidth}{!}{%
\begin{tabular}{@{}l cccccc cccccc@{}}
\toprule
& \multicolumn{6}{c}{\textbf{RORD-Val}}
& \multicolumn{6}{c}{\textbf{OBER-Test}} \\
\cmidrule(lr){2-7}
\cmidrule(lr){8-13}
\textbf{Method}
& {\scriptsize \textbf{FID $\downarrow$}}
& {\scriptsize \textbf{localFID $\downarrow$}}
& {\scriptsize \textbf{fMSE $\downarrow$}}
& {\scriptsize \textbf{PSNR$_{\text{FG}}$ $\uparrow$}}
& {\scriptsize \textbf{LPIPS $\downarrow$}}
& {\scriptsize \textbf{CLIP $\downarrow$}}
& {\scriptsize \textbf{FID $\downarrow$}}
& {\scriptsize \textbf{localFID $\downarrow$}}
& {\scriptsize \textbf{fMSE $\downarrow$}}
& {\scriptsize \textbf{PSNR$_{\text{FG}}$ $\uparrow$}}
& {\scriptsize \textbf{LPIPS $\downarrow$}}
& {\scriptsize \textbf{CLIP $\downarrow$}} \\
\midrule

PowerPaint (ECCV'24)
& 53.81  & 65.34  & 2508.69 & 15.84 & 0.2346 & 0.0715
& 67.37  & 90.98  & 1916.73 & 18.62 & 0.1243 & 0.0409 \\

SDXL (ICLR'24)
& 67.18  & 79.72  & 2947.48 & 14.74 & 0.2308 & 0.1101
& 75.44  & 109.50 & 2573.60 & 16.15 & 0.1296 & 0.0579 \\

OmniEraser (arXiv'25)
& 46.80  & 51.65  & 1247.13 & 18.62 & 0.1962 & 0.0536
& 54.70  & 54.86  & 883.91  & 21.00 & 0.1783 & \underline{0.0142} \\

GeoRemover (NeurIPS'25)
& \underline{27.77} & \underline{42.28} & \underline{792.00} & \textbf{21.18} & \textbf{0.1009} & \underline{0.0320}
& 50.08  & 50.80  & 769.53  & 22.12 & 0.1976 & 0.0670 \\

LBM (ICCV'25)
& 43.01  & 60.37  & 1301.02 & 18.45 & 0.1768 & 0.0470
& 29.13  & 54.20  & \underline{477.60} & \underline{23.75} & 0.0824 & 0.0171 \\

FLAIR (NeurIPS'25)
& 55.26  & 80.10  & 1846.13 & 17.16 & 0.2488 & 0.0660
& 40.64  & 70.27  & 686.14  & 22.38 & 0.2071 & 0.0244 \\

ObjectClear (CVPR'26)
& 33.61  & 50.92  & 1368.08 & 18.65 & 0.1238 & 0.0360
& \textbf{18.86} & \textbf{44.36} & 588.05 & \underline{23.75} & \textbf{0.0342} & \textbf{0.0103} \\

\midrule
\textbf{Ours}
& \textbf{26.92} & \textbf{36.18} & \textbf{729.24} & \underline{21.13} & \underline{0.1211} & \textbf{0.0280}
& \underline{24.39} & \underline{49.67} & \textbf{460.58} & \textbf{23.93} & \underline{0.0748} & 0.0152 \\

\bottomrule
\end{tabular}}

\label{Table:comparison_combined}

\end{table*}

\begin{figure*}[t]
  \centering
  \setlength{\tabcolsep}{0.5pt}
  \renewcommand{\arraystretch}{1.0}
  \def\cellw{0.235\textwidth}
  \def\cellh{0.132\textwidth}
  \begin{tabular}{@{}c@{\hspace{1.5pt}}cccc@{}}

    \rotatebox{90}{\footnotesize \makebox[\cellh][c]{Input+Mask}} &
    \includegraphics[width=\cellw]{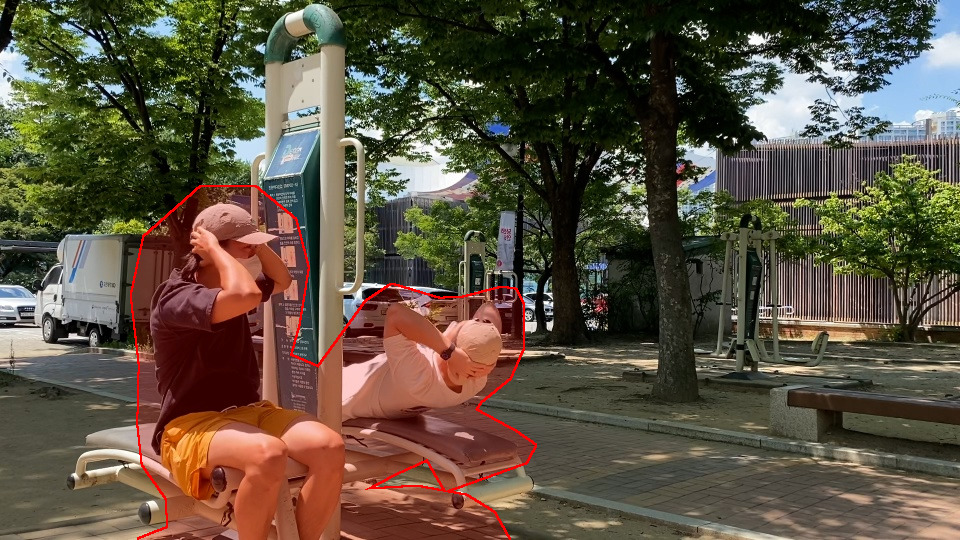} &
    \includegraphics[width=\cellw]{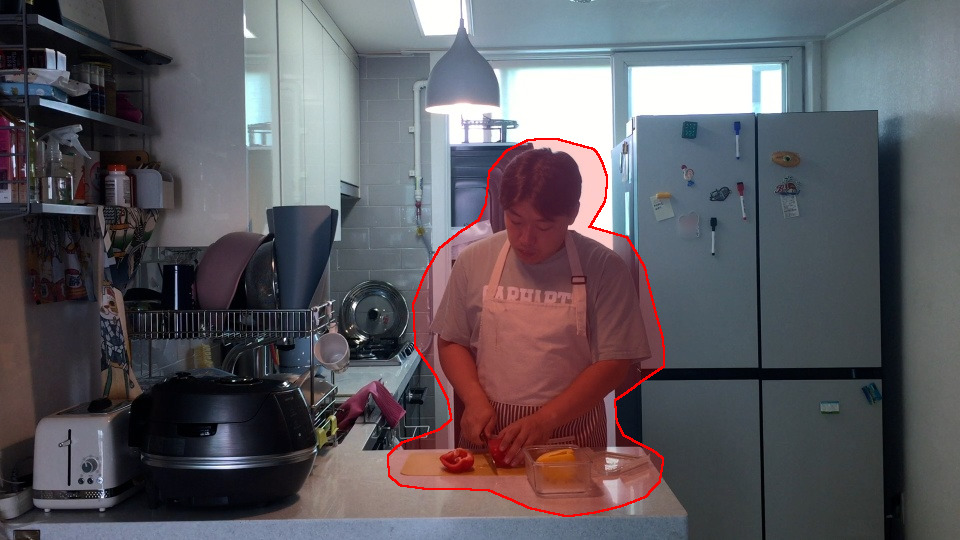} &
    \includegraphics[width=\cellw]{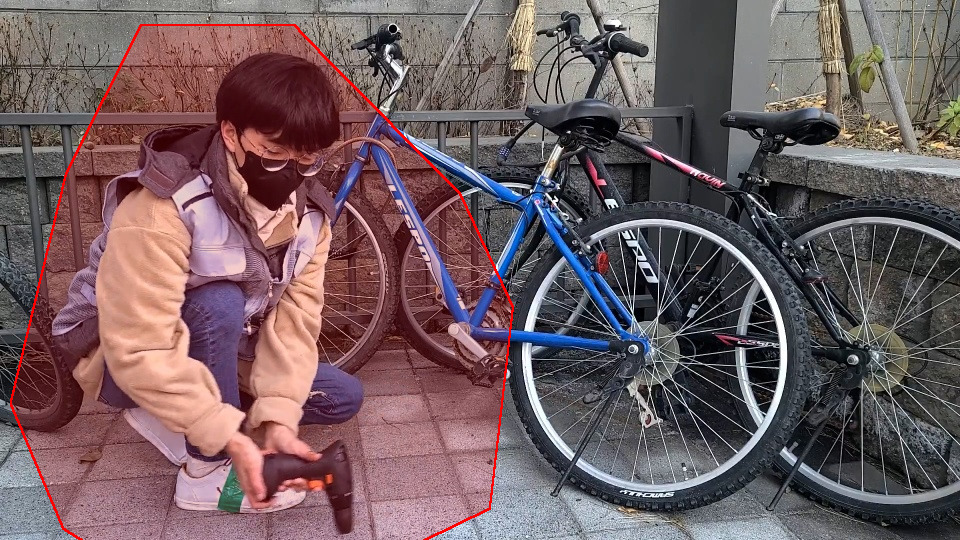} &
    \includegraphics[width=\cellw]{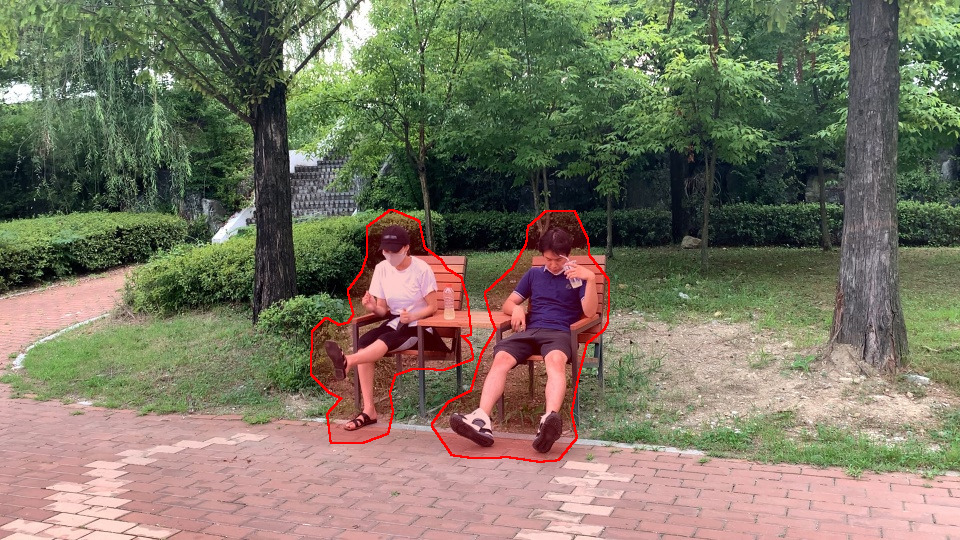} \\[-1.5pt]

    \rotatebox{90}{\footnotesize \makebox[\cellh][c]{PowerPaint}} &
    \includegraphics[width=\cellw]{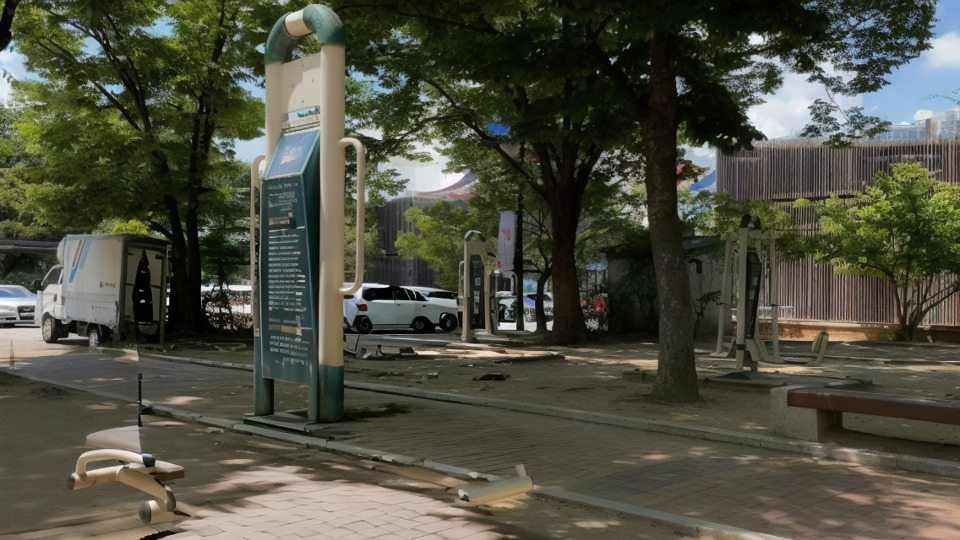} &
    \includegraphics[width=\cellw]{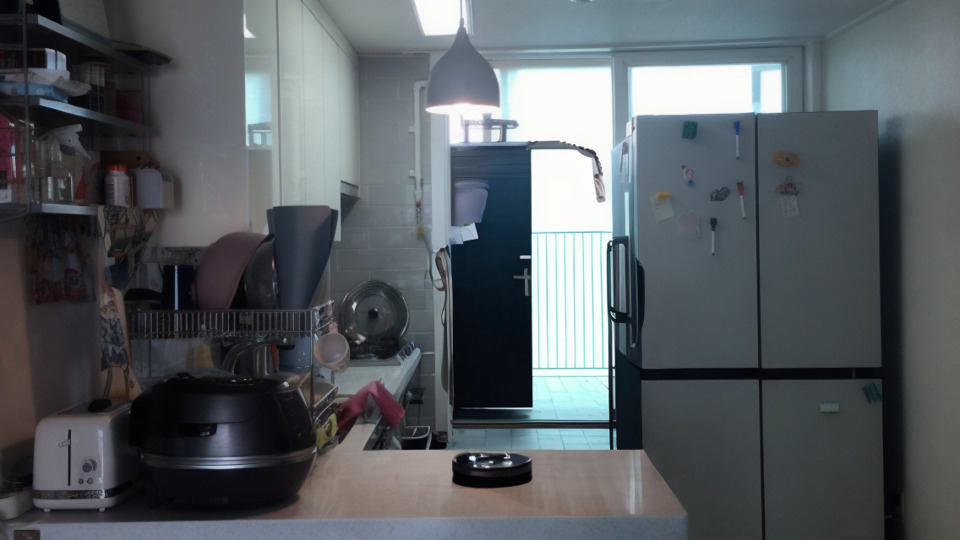} &
    \includegraphics[width=\cellw]{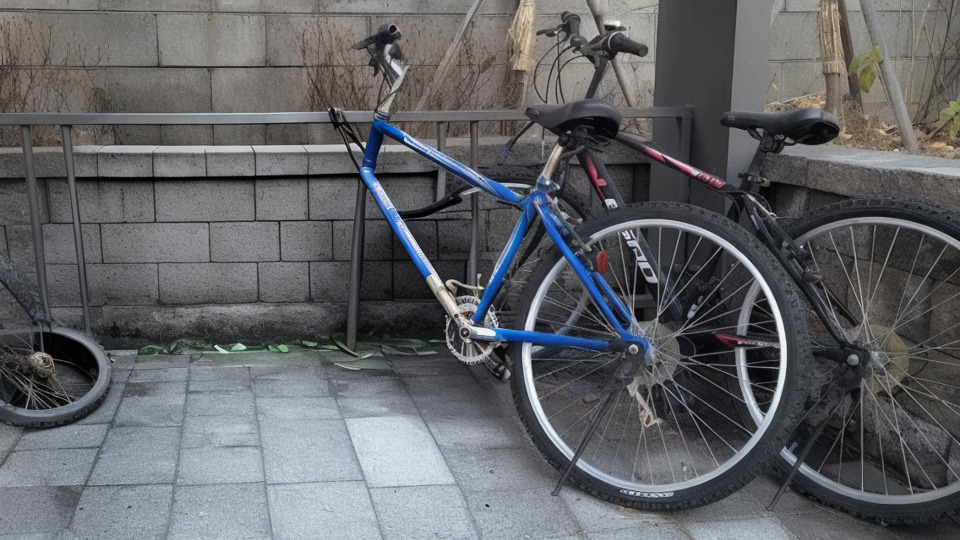} &
    \includegraphics[width=\cellw]{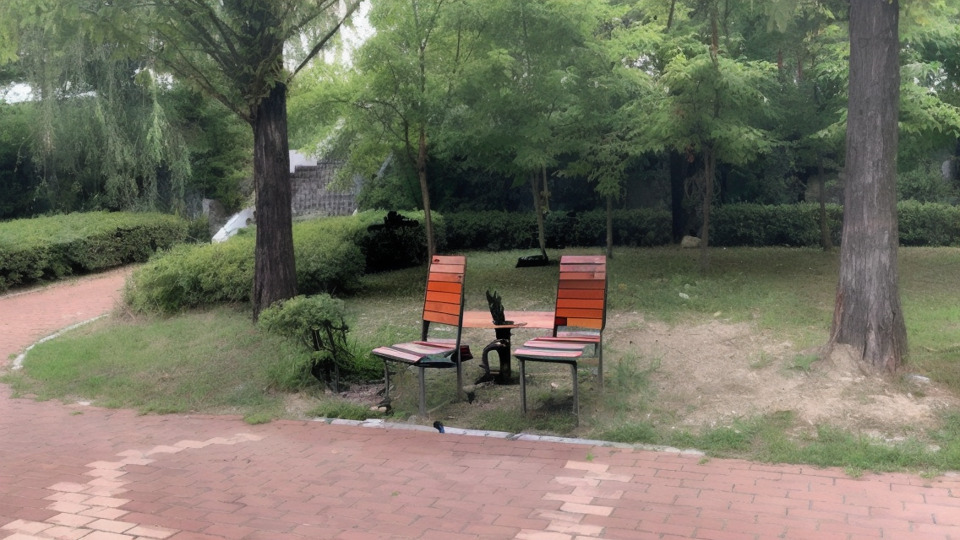} \\[-1.5pt]

    \rotatebox{90}{\footnotesize \makebox[\cellh][c]{OmniEraser}} &
    \includegraphics[width=\cellw]{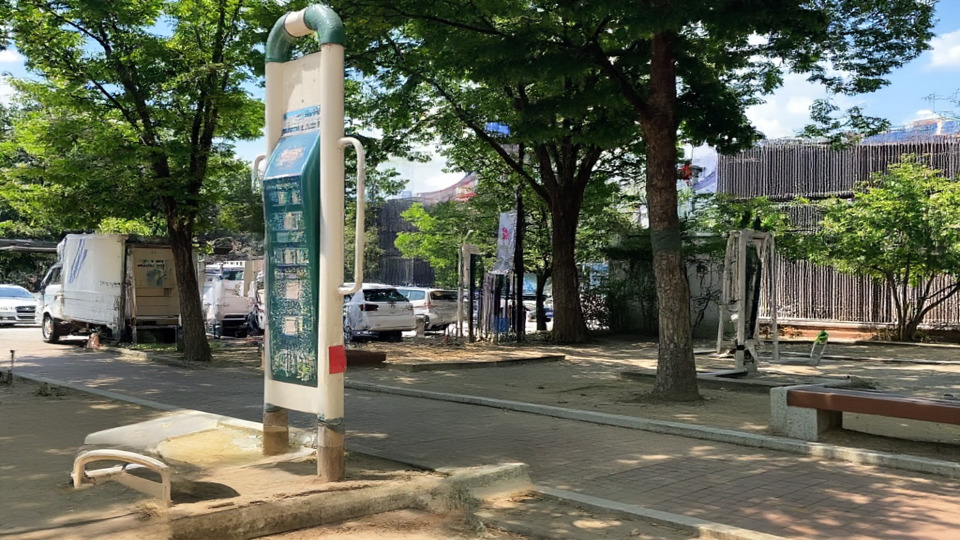} &
    \includegraphics[width=\cellw]{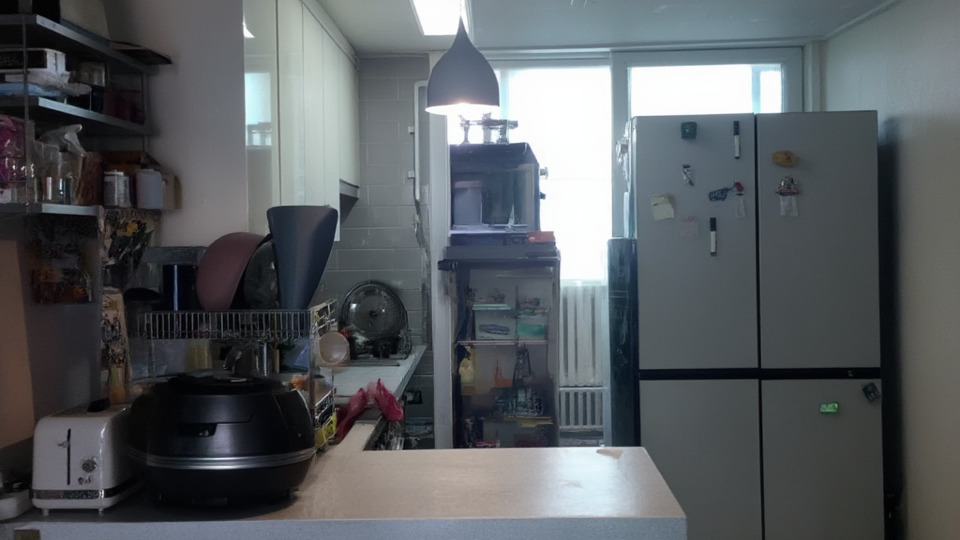} &
    \includegraphics[width=\cellw]{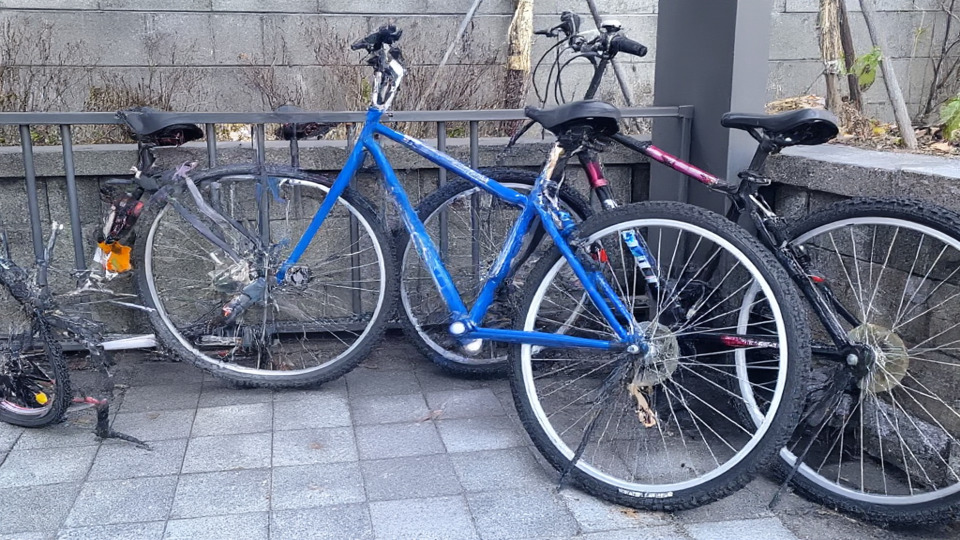} &
    \includegraphics[width=\cellw]{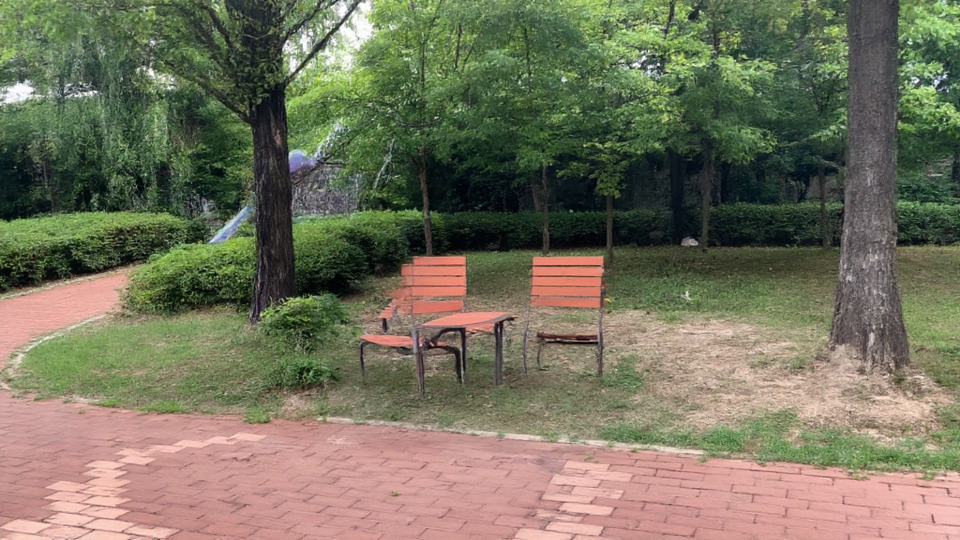} \\[-1.5pt]

    \rotatebox{90}{\footnotesize \makebox[\cellh][c]{GeoRemover}} &
    \includegraphics[width=\cellw]{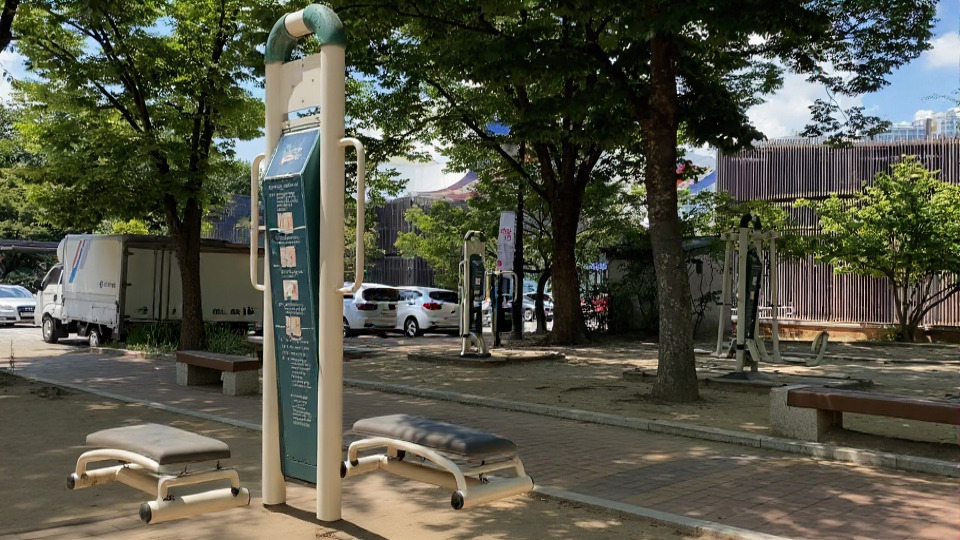} &
    \includegraphics[width=\cellw]{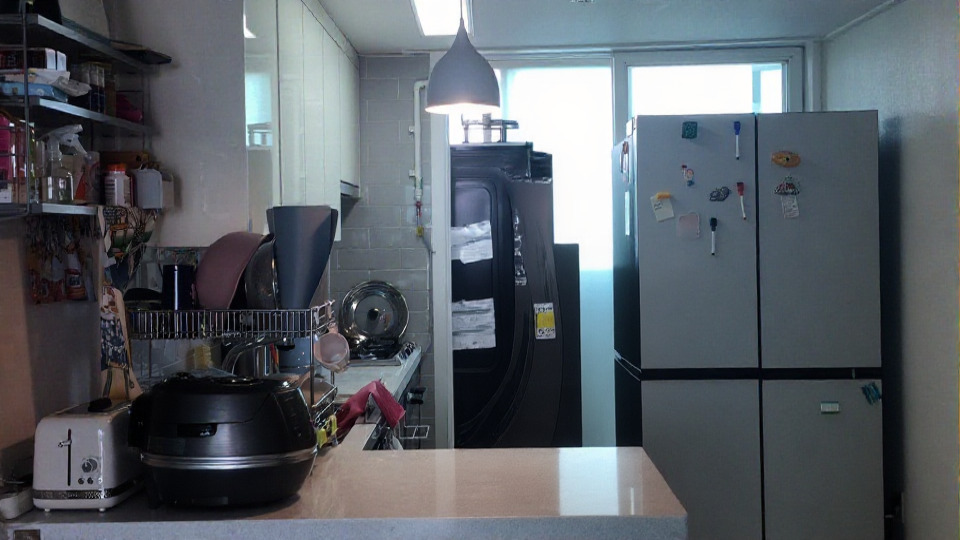} &
    \includegraphics[width=\cellw]{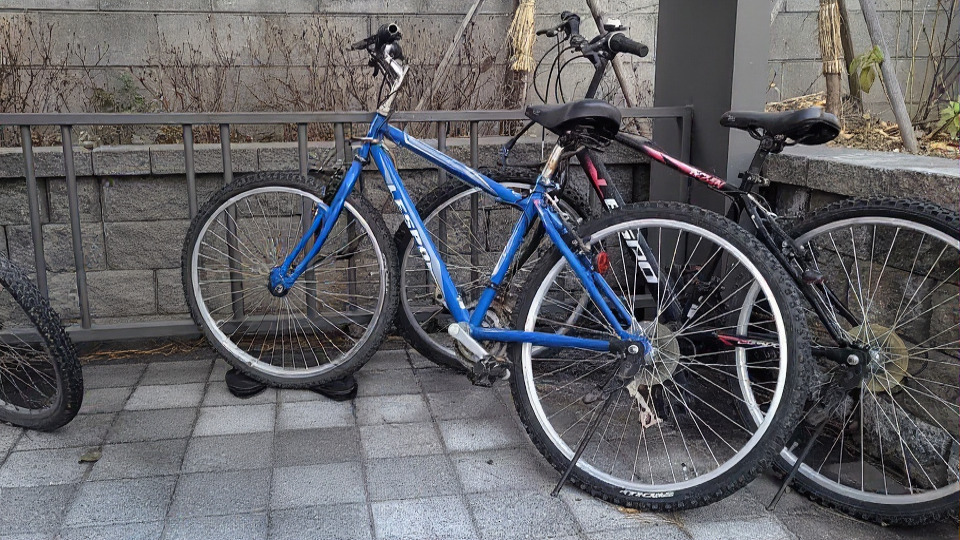} &
    \includegraphics[width=\cellw]{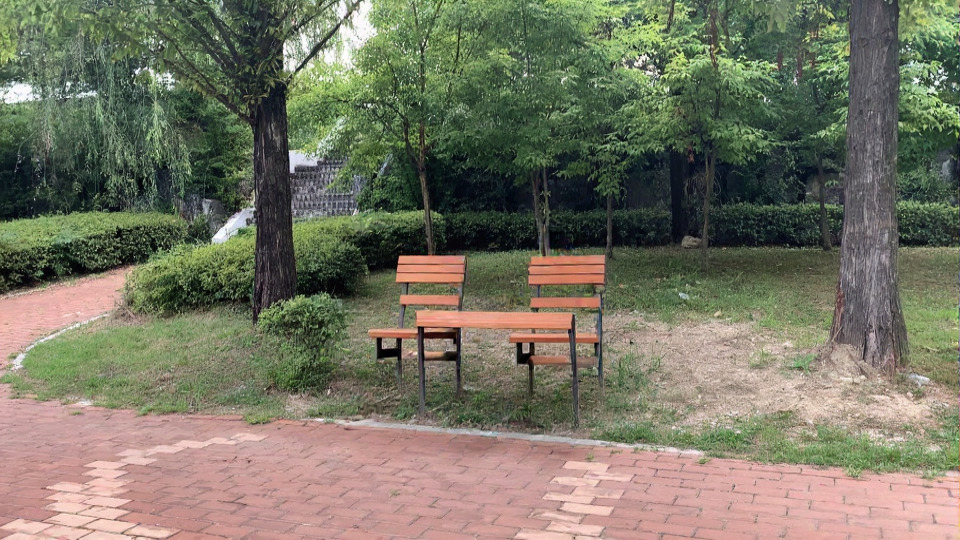} \\[-1.5pt]

    \rotatebox{90}{\footnotesize \makebox[\cellh][c]{ObjectClear}} &
    \includegraphics[width=\cellw]{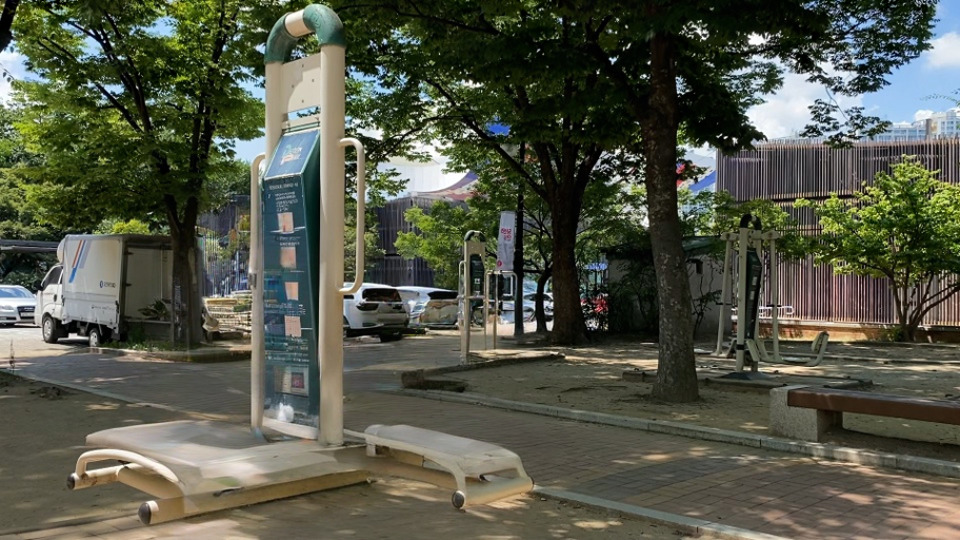} &
    \includegraphics[width=\cellw]{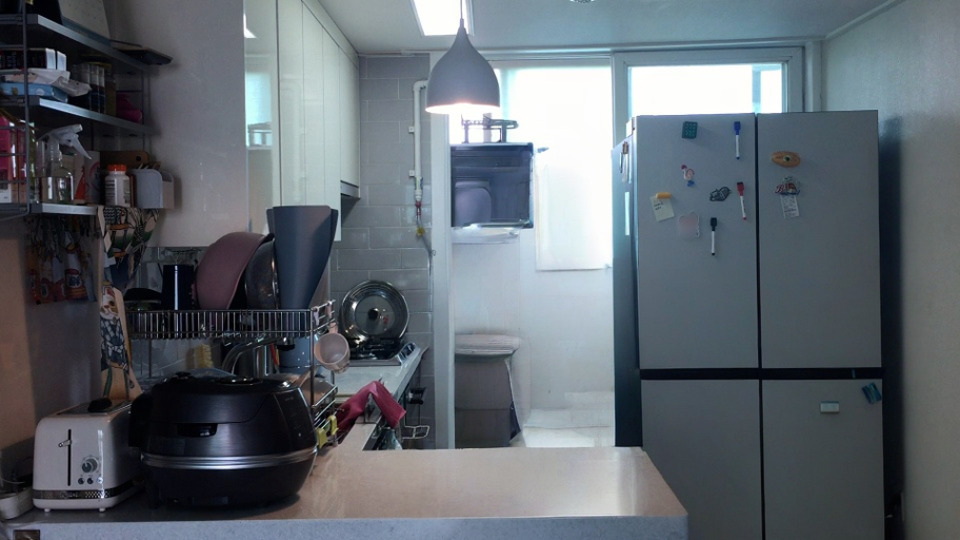} &
    \includegraphics[width=\cellw]{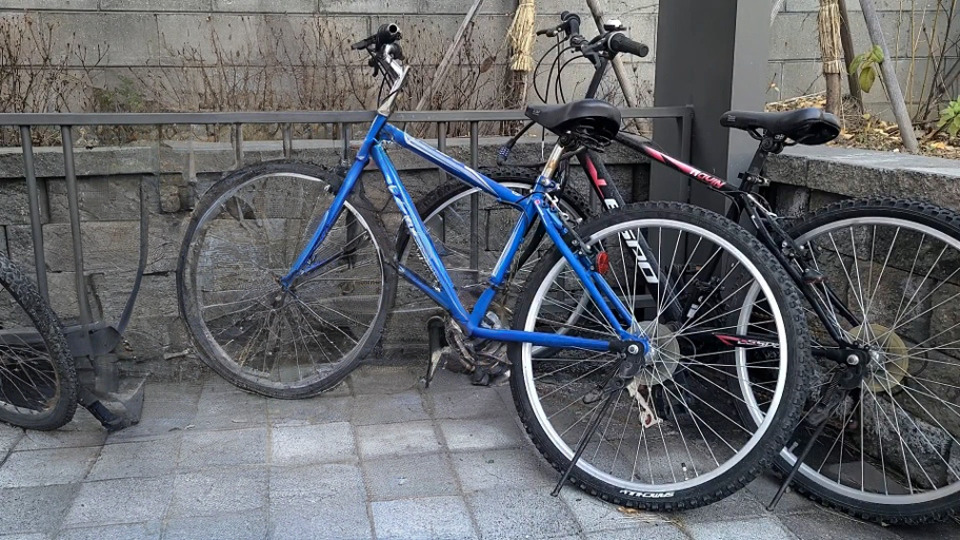} &
    \includegraphics[width=\cellw]{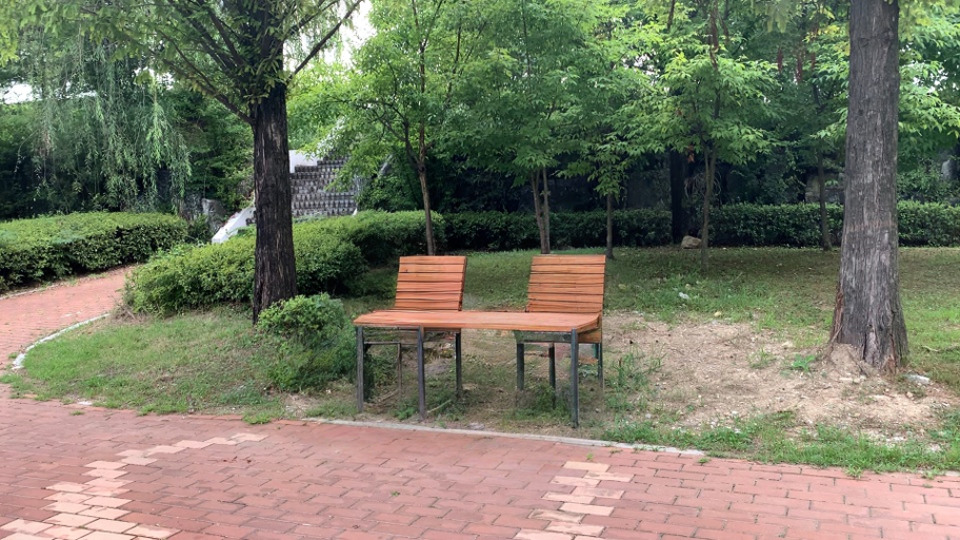} \\[-1.5pt]

    \rotatebox{90}{\footnotesize \makebox[\cellh][c]{\textbf{SERO (Ours)}}} &
    \includegraphics[width=\cellw]{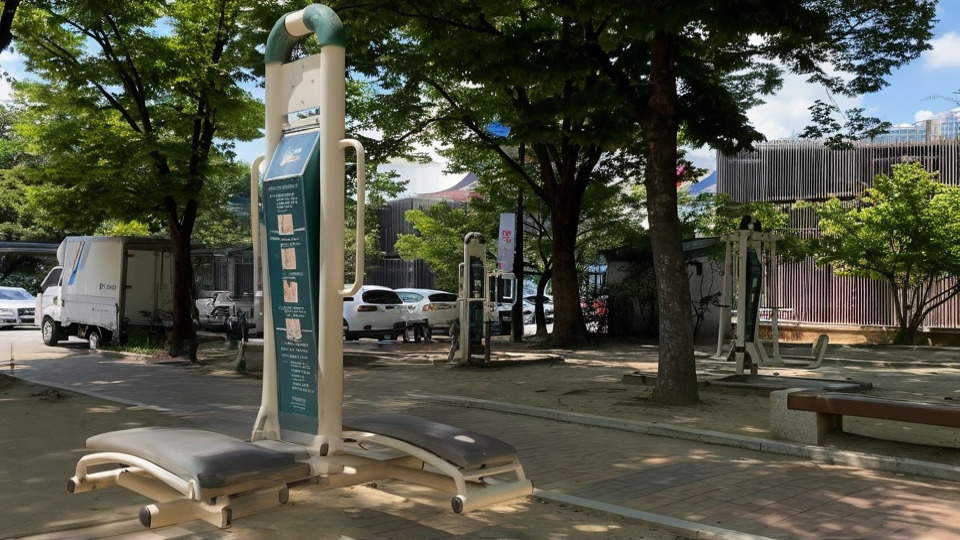} &
    \includegraphics[width=\cellw]{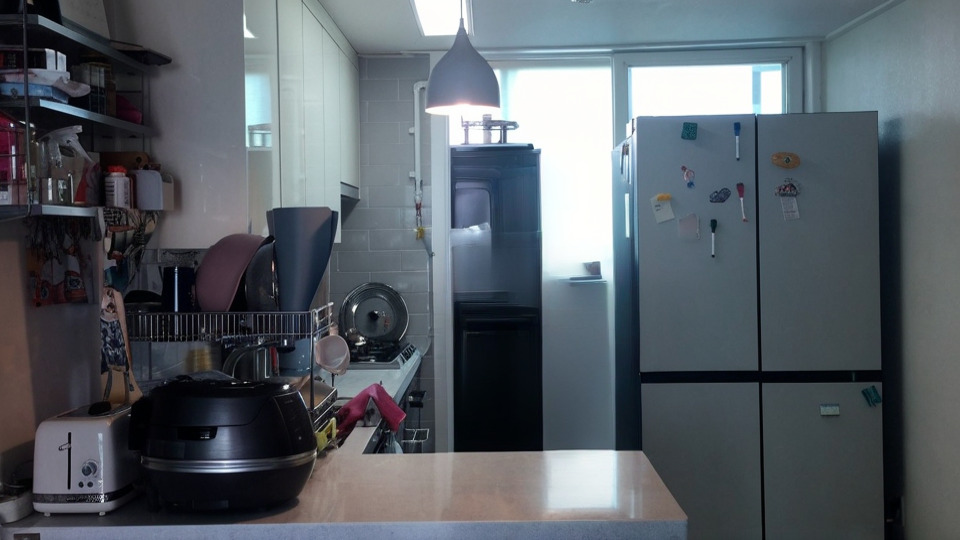} &
    \includegraphics[width=\cellw]{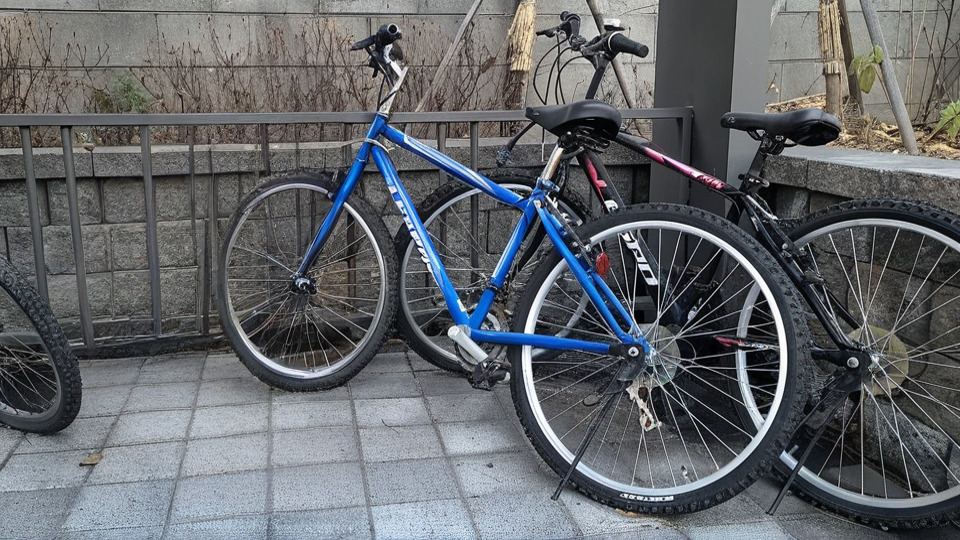} &
    \includegraphics[width=\cellw]{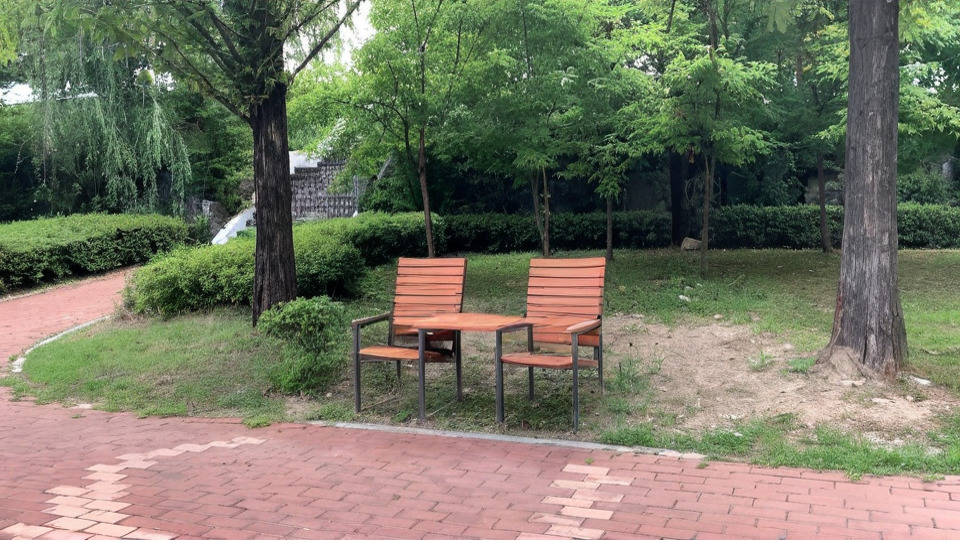} \\[-1.5pt]

    \rotatebox{90}{\footnotesize \makebox[\cellh][c]{GT}} &
    \includegraphics[width=\cellw]{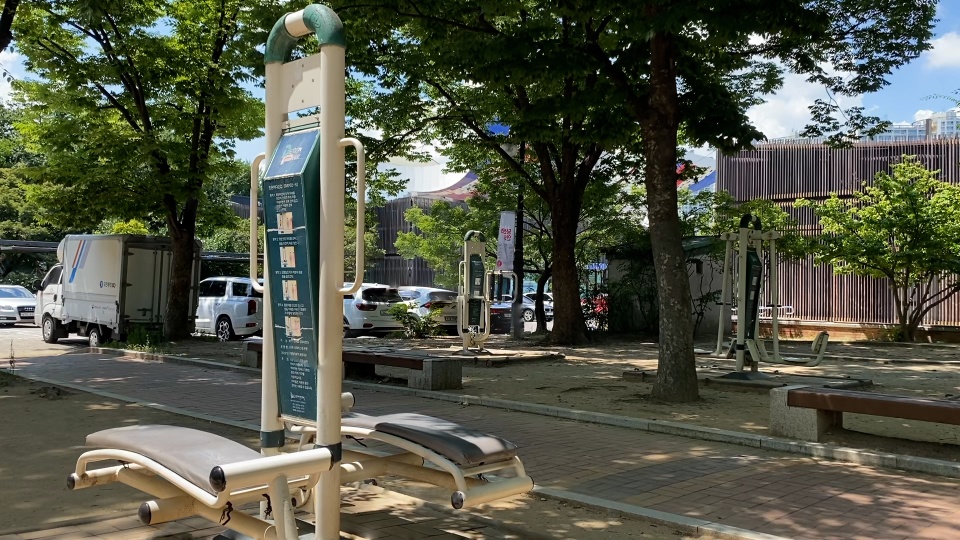} &
    \includegraphics[width=\cellw]{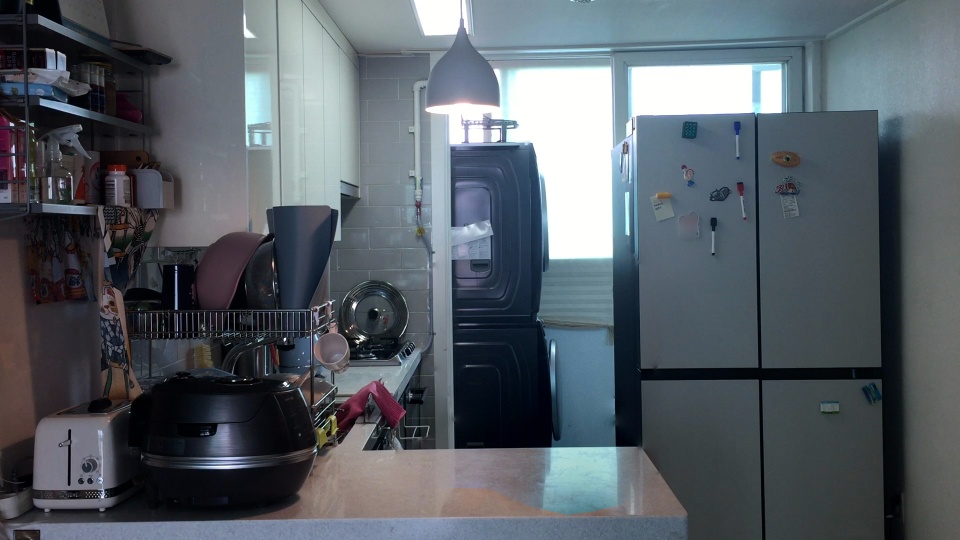} &
    \includegraphics[width=\cellw]{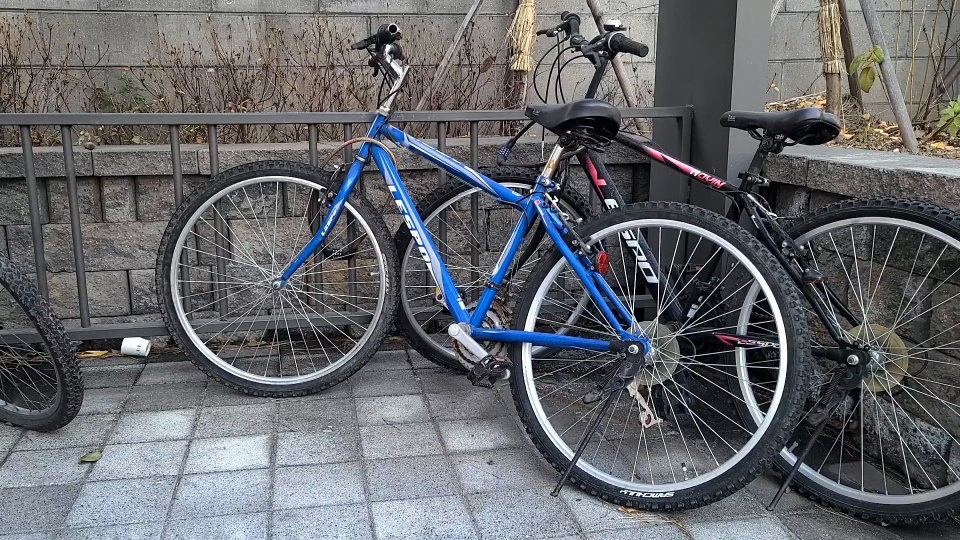} &
    \includegraphics[width=\cellw]{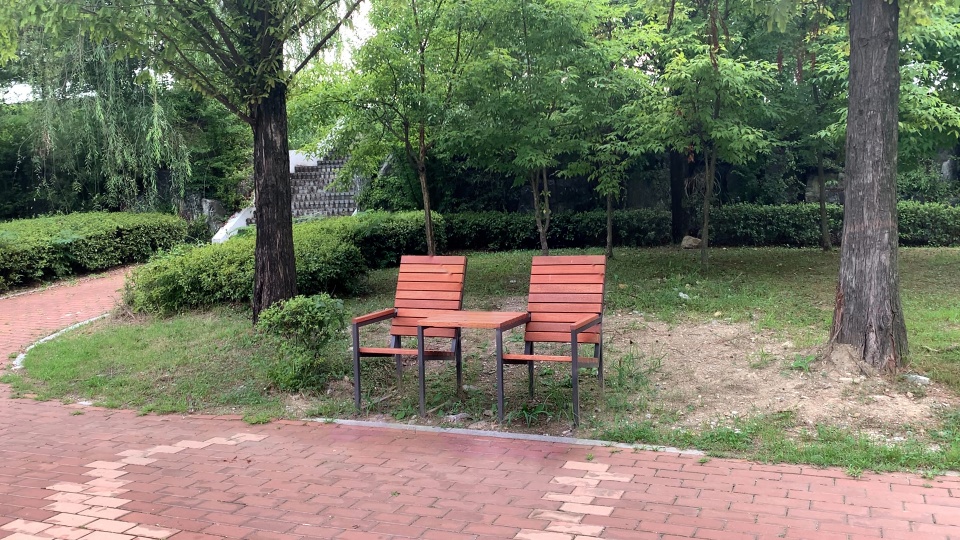} \\
  \end{tabular}
  \caption{\textbf{Qualitative comparison on RORD-Val.} Each row corresponds to
    one method and each column to one scene. Compared with PowerPaint,
    OmniEraser, GeoRemover, and ObjectClear, SERO removes the target object
    while producing the most semantically coherent and artifact-free background
    completion, closely matching the ground truth.}
  \Description{Four RORD-Val scenes arranged as columns, with one row per method: the masked input, four prior methods, SERO, and the ground-truth background.}
  \label{fig:qual_rord}
\end{figure*}

\begin{figure*}[t]
  \centering
  \setlength{\tabcolsep}{0.5pt}
  \renewcommand{\arraystretch}{1.0}
  \def\cellwo{0.137\textwidth}
  \begin{tabular}{@{}ccccccc@{}}
    \footnotesize Input+Mask & \footnotesize PowerPaint &
    \footnotesize OmniEraser & \footnotesize GeoRemover &
    \footnotesize ObjectClear & \footnotesize \textbf{SERO (Ours)} &
    \footnotesize GT \\
    \includegraphics[width=\cellwo]{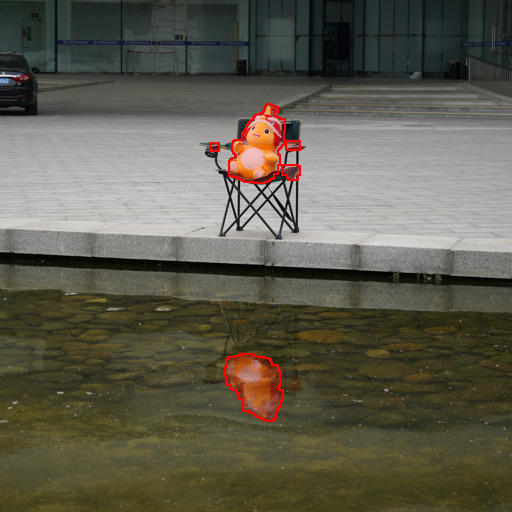} &
    \includegraphics[width=\cellwo]{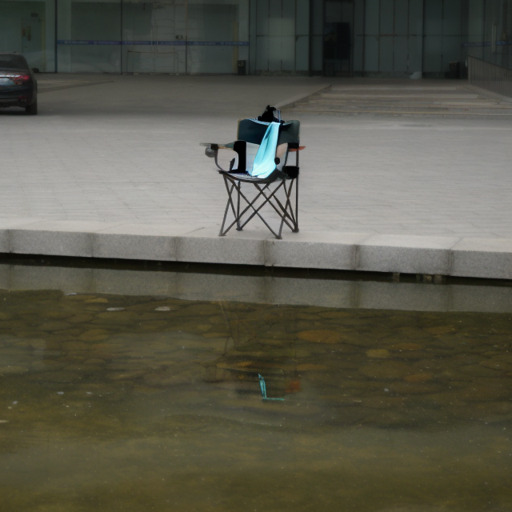} &
    \includegraphics[width=\cellwo]{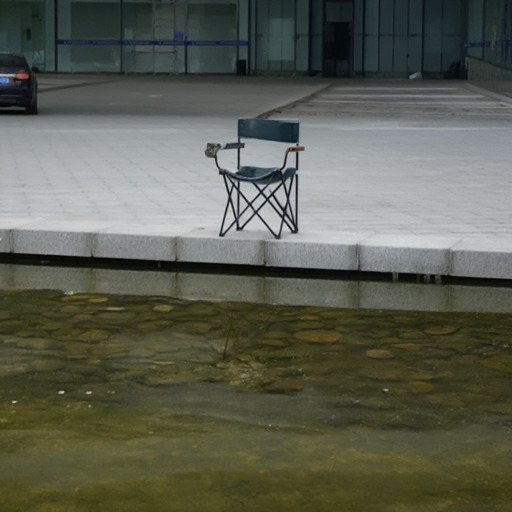} &
    \includegraphics[width=\cellwo]{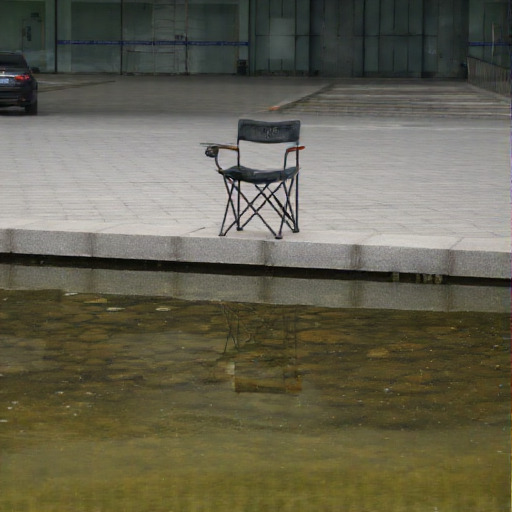} &
    \includegraphics[width=\cellwo]{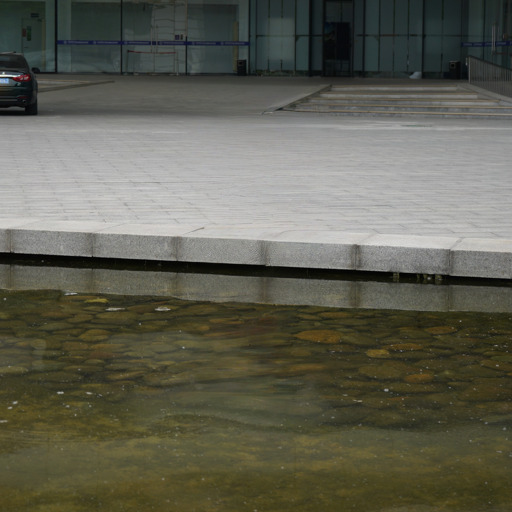} &
    \includegraphics[width=\cellwo]{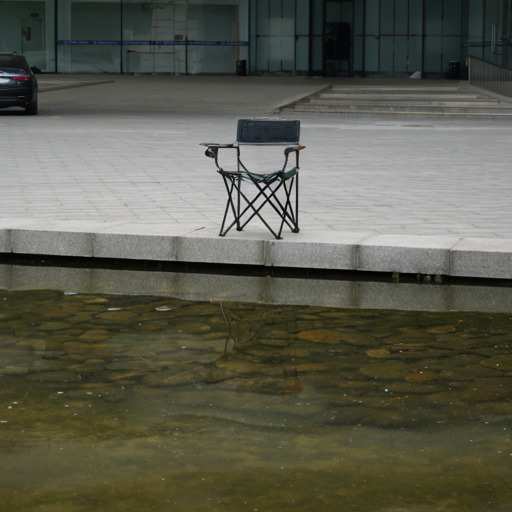} &
    \includegraphics[width=\cellwo]{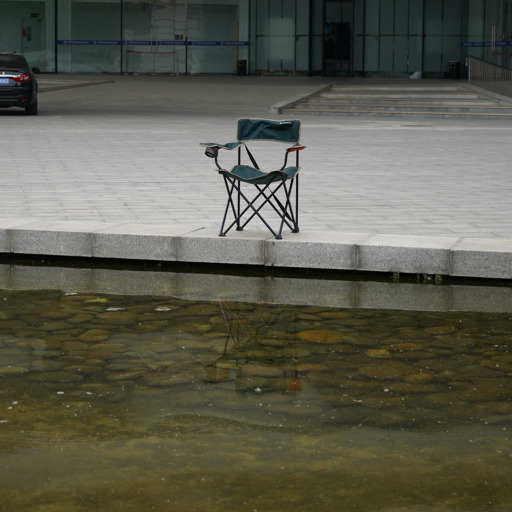} \\[-1.5pt]
    \includegraphics[width=\cellwo]{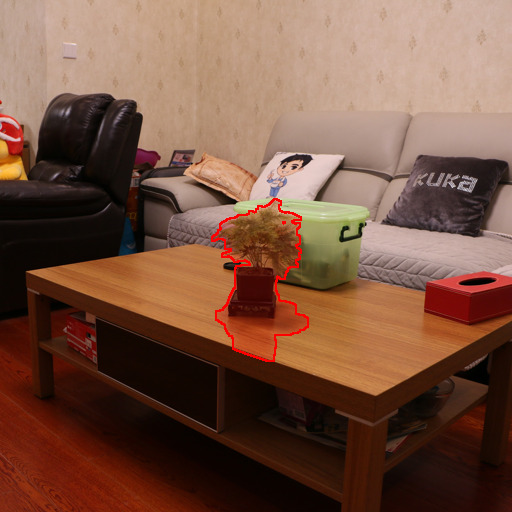} &
    \includegraphics[width=\cellwo]{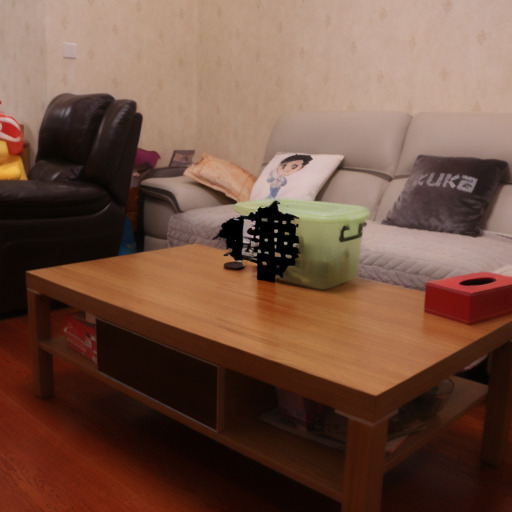} &
    \includegraphics[width=\cellwo]{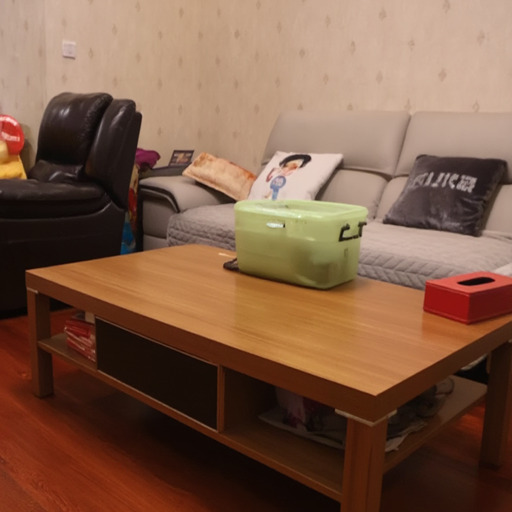} &
    \includegraphics[width=\cellwo]{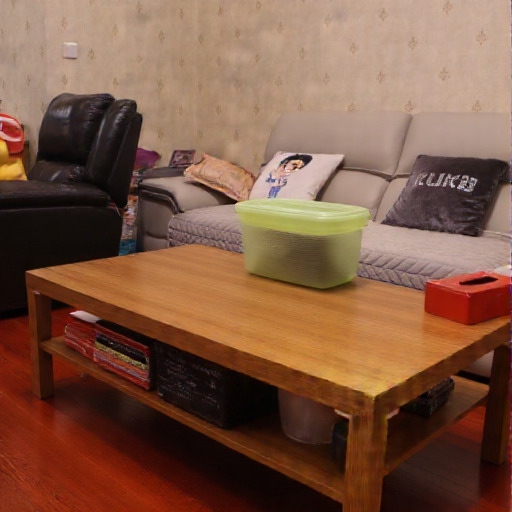} &
    \includegraphics[width=\cellwo]{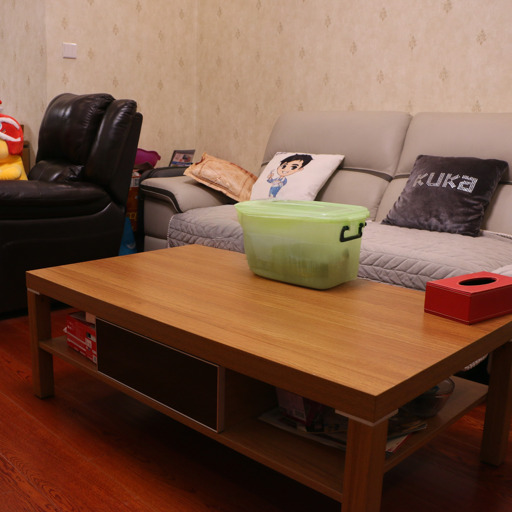} &
    \includegraphics[width=\cellwo]{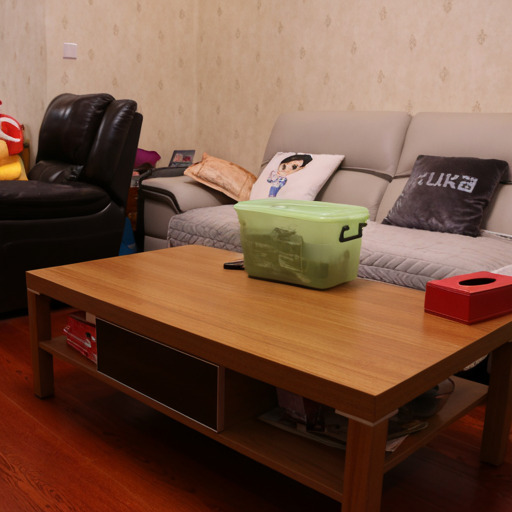} &
    \includegraphics[width=\cellwo]{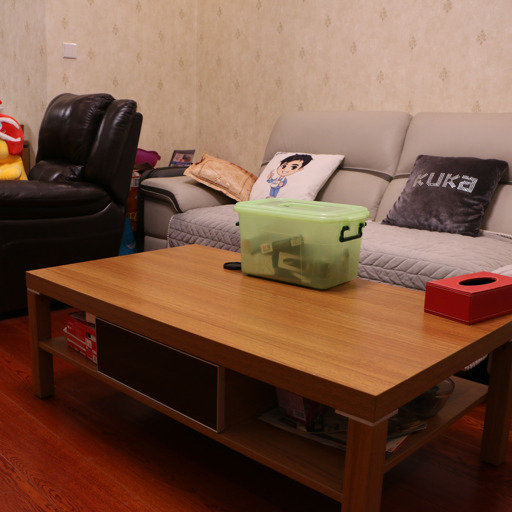} \\[-1.5pt]
    \includegraphics[width=\cellwo]{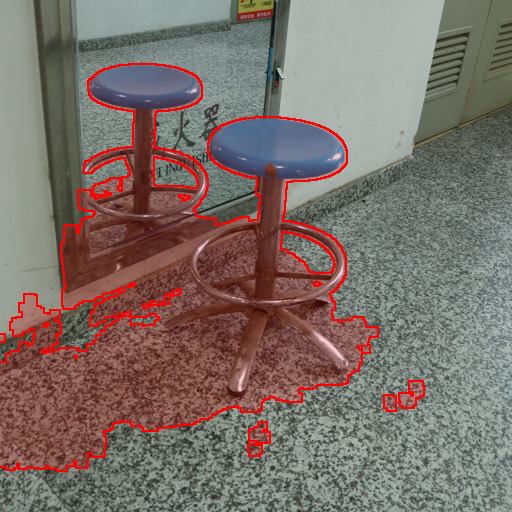} &
    \includegraphics[width=\cellwo]{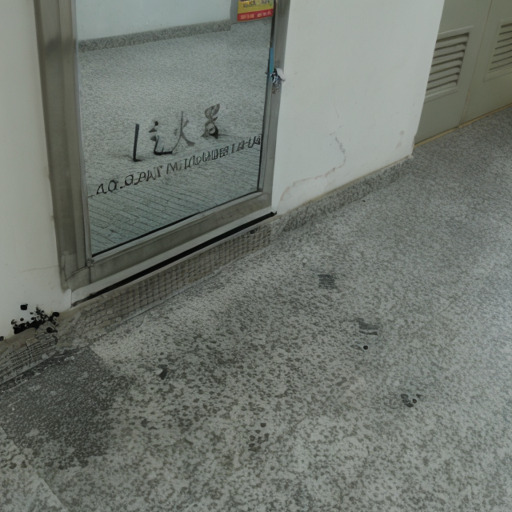} &
    \includegraphics[width=\cellwo]{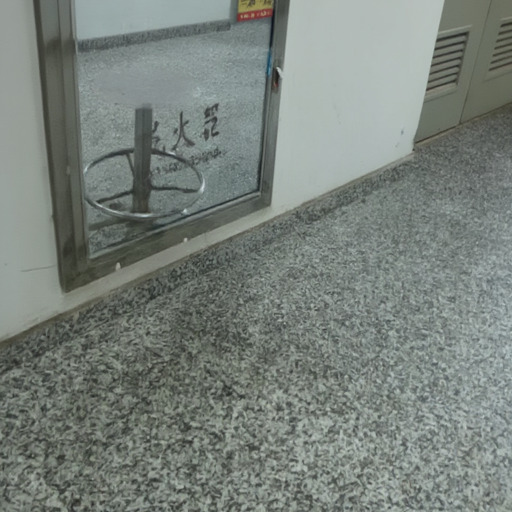} &
    \includegraphics[width=\cellwo]{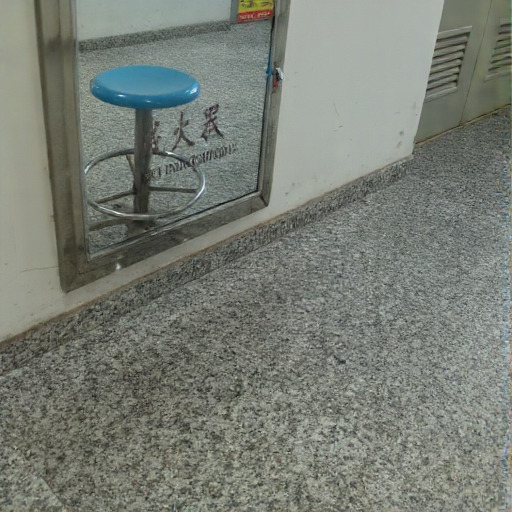} &
    \includegraphics[width=\cellwo]{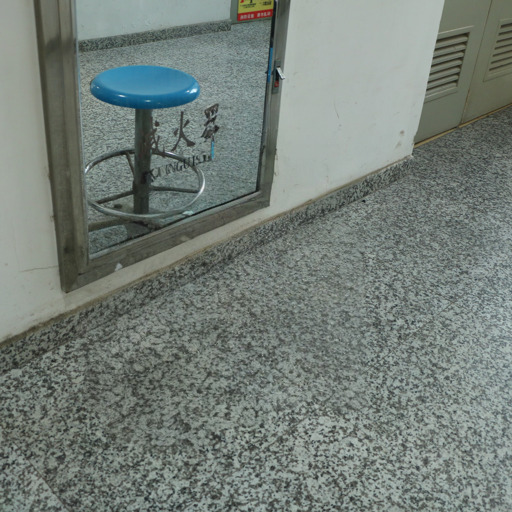} &
    \includegraphics[width=\cellwo]{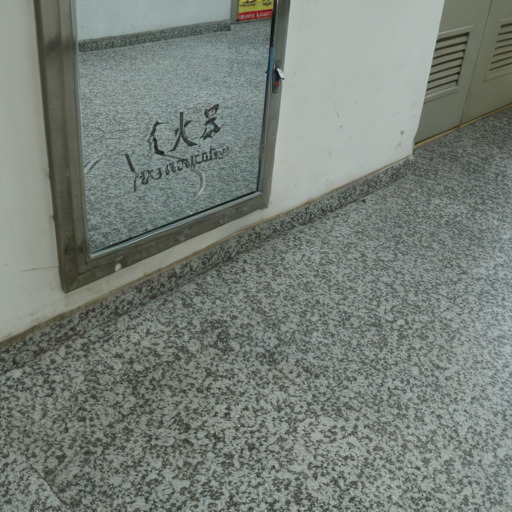} &
    \includegraphics[width=\cellwo]{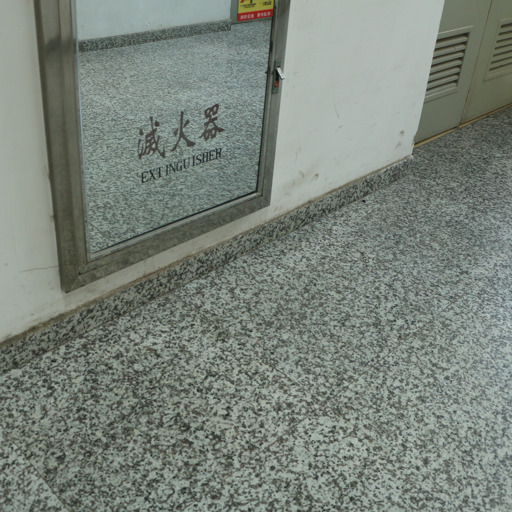} \\
  \end{tabular}
  \caption{\textbf{Qualitative comparison on OBER-Test.} Each column
    corresponds to one method and each row to one scene. In addition to the
    object itself, SERO removes the associated visual effects, such as the
    reflection in the first row, while preserving the surrounding scene; prior
    methods either leave residual traces of the effect or alter content outside
    the mask.}
  \Description{Three OBER-Test scenes arranged as rows, with one column per method: the masked input, four prior methods, SERO, and the ground-truth background.}
  \label{fig:qual_ober}
\end{figure*}

\subsection{Training Objective}
\label{sec:train}
We optimise the standard LBM velocity-matching objective:
\begin{equation}
  \mathcal{L}_{\text{LBM}} =
  \mathbb{E}_{t,\,\varepsilon}\!\left[
    \left\|
    \frac{z_1 - z_t}{1-t}
    -
    v_\theta(z_t,\,t,\,c)
    \right\|^2_{\Omega}
  \right],
  \label{eq:lbm-loss}
\end{equation}
where $\Omega$ denotes the valid training region derived from the removal mask.

To improve perceptual quality, we further add a multi-scale LPIPS loss~\cite{zhang2018unreasonable}, following the common use of multi-scale perceptual supervision in image synthesis~\cite{wang2018high}:
\begin{equation}
  \mathcal{L}_{\text{perc}} =
  \sum_{s \in \mathcal{S}} w_s \,
  \mathrm{LPIPS}\!\left(\hat{x}_{1,s}, x_{1,s}, \Omega_s\right),
  \label{eq:lpips-loss}
\end{equation}
where $\mathcal{S}=\{1.0,0.5,0.25\}$ and $w=\{0.5,0.3,0.2\}$. This multi-scale perceptual term noticeably improves reconstruction quality over the base LBM objective, especially in structural consistency and local details.

The final loss is
\begin{equation}
  \mathcal{L}
  =
  \mathcal{L}_{\text{LBM}}
  +
  \lambda_{\text{perc}}\mathcal{L}_{\text{perc}}.
  \label{eq:total-loss}
\end{equation}
While training, only the diffusion backbone and decoder adapter are updated, while the context encoder and predictor remain frozen.
\paragraph{Inference.}
At test time, SERO takes the input image $x_0$ together with a removal mask $m$ as input. Following~\cite{chadebec2025lbm}, the target latent can be efficiently estimated from an intermediate latent state $z_t$ as
\begin{equation}
\hat{z}_1 = z_t + (1-t)\,v_\theta(z_t,t,c),
\qquad
\hat{x}_1 = \mathcal{D}(\hat{z}_1),
\label{eq:z1hat}
\end{equation}
where $\hat{z}_1$ denotes the predicted target latent,
$\mathcal{D}$ is the frozen SDXL VAE decoder, and
$\hat{x}_1$ is the reconstructed output image.

When a removal mask is unavailable, SERO also supports prompt-driven inference. Specifically, a user-provided text prompt describing the target object is passed, together with the input image, to the prompt-based segmentation method of Ren \emph{et al.}~\cite{ren2024grounded} to automatically generate the removal mask. The generated mask is then used as $m$, after which the standard inference pipeline proceeds unchanged (Fig ~\ref{fig:arch}b).

\section{Experiments}

In this section, we perform extensive experiments to analyse and evaluate our network, SERO. Sec.~\ref{sec:setting} describes the experimental settings, Sec.~\ref{sec:sota_comparison} compares SERO with existing methods, and Sec.~\ref{sec:ablation} illustrates our ablations. 
\subsection{Setting}
\label{sec:setting}

\subsubsection{Datasets} We train and evaluate on two object removal benchmarks RORD~\cite{Sagong_2022_BMVC} and {OBER}~\cite{zhao2026objectclear}. RORD is a large-scale real-world object removal dataset
containing training and validation images.
Each sample includes an input image, a clean ground-truth image, and a single coarse
mask covering both the object and its effect. For evaluation, from the validation split, we select one image per scene as in \cite{zhao2026objectclear}, thereby avoiding duplicates, resulting in 343 samples in the RORD-Val dataset. OBER is an object and effect removal dataset consisting of training and test images (163 samples in OBER-Test split).
Each sample includes an input image with one or more objects, a clean ground-truth image, a precise object mask, and a separate object-effect mask that covers both the
object and its associated effect (\emph{e.g. } shadows, reflections).
We use the object-effect mask as the removal target, as it encompasses the full
extent of the visual artefact to be removed.
We train on the OBER training set and RORD training set and report
quantitative results on the OBER-Test and RORD-Val set separately.

\subsubsection{Implementation Details}
We trained our model for 212k iterations with the OBER training dataset and 172k iterations with RORD training dataset on one NVIDIA H200 GPU using a batch size of 16 and a learning rate of 4e-5. We used a pre-trained I-JEPA ViT-H/14 model \cite{assran2023self} at an input resolution of $224 \times 224$ as a frozen semantic prior. This corresponds to a patch size of $14 \times 14$ and a resulting token grid of $16 \times 16$. We compare SERO with two groups of baselines: (1)~\textit{general image inpainting methods}: SDXL-Inpainting~\cite{podell2024sdxl}, PowerPaint~\cite{zhuang2024powerpaint}, FLAIR~\cite{erbach2026solving} and LBM~\cite{chadebec2025lbm}, (2)~\textit{object and effect removal methods}: OmniEraser~\cite{wei2025omnieraser}, GeoRemover~\cite{zhu2025georemover}, and ObjectClear~\cite{zhao2026objectclear}. Since pretrained object-removal checkpoints for LBM were not publicly available, we trained the model ourselves on the same data and evaluated it on both benchmarks.
General inpainting methods serve as general-purpose baselines for scene completion quality but are not specifically designed to address the full object-effect removal problem; dedicated removal methods explicitly target the complete removal of both objects and their associated visual effects.

\subsubsection{Evaluation Metrics} We consider three aspects: image generation quality, masked-region fidelity, and perceptual quality. We assess image generation quality using the Fréchet Inception Distance (FID)~\cite{heusel2017gans}. Since our task involves masked region manipulation, which occupies a small portion of the whole image, global FID alone cannot well reflect the generation quality in this region.  Therefore, for masked-region fidelity, we report local FID~\cite{xie2023smartbrush}, foreground MSE (fMSE), and PSNR-FG~\cite{hore2010image}, each of which are computed only within the masked region. Finally, we use two perceptual metrics: LPIPS~\cite{zhang2018unreasonable} and CLIP feature distance~\cite{radford2021learning}.
\subsection{Comparison with State-of-the-Art Methods}
\label{sec:sota_comparison}
\subsubsection{Quantitative Evaluation} As shown in Table~\ref{Table:comparison_combined}, our method achieves the strongest overall performance on RORD-Val. It obtains the best FID, localFID, fMSE, and CLIP distance, with scores of 26.92, 36.18, 729.24, and 0.0280, respectively. These results show that our method improves both global image realism and local consistency in the edited region. In the mask area, which corresponds to the region that must be regenerated by the model, our method achieves the lowest fMSE and remains highly competitive in PSNR$_{\text{FG}}$. It also obtains the second-best LPIPS, further indicating strong perceptual quality. Compared with LBM \cite{chadebec2025lbm}, SERO augments the mask-based spatial conditioning with a predictive semantic prior through concatenation.

We further evaluate on OBER-Test, which includes object removal together with the associated effect removal. Although several baselines are specifically designed for this setting, our method achieves the best mask-region reconstruction accuracy, with the lowest fMSE of 460.58 and the highest PSNR$_{\text{FG}}$ of 23.93. Beyond the masked region, it also maintains strong global and local realism, ranking second in FID, localFID, and LPIPS.  Importantly, GeoRemover~\cite{zhu2025georemover} and ObjectClear~\cite{zhao2026objectclear} require 100 and 20 inference steps, respectively, whereas our model achieves high-quality object-and-effect removal in a single step. This demonstrates that our approach not only produces accurate reconstructions in the generated mask region, but also offers a more efficient balance between visual quality and inference cost.

\subsubsection{Qualitative Evaluation}
Figs.~\ref{fig:qual_rord} and~\ref{fig:qual_ober} show qualitative comparisons with representative object removal methods on RORD-Val and OBER-Test, respectively. Since RORD-Val images are wide-aspect, we arrange that comparison with one method per row and one scene per column, and use the transposed layout for the square OBER-Test images. Existing methods often modify regions outside the target mask or introduce artifacts near the removed object. For example, PowerPaint frequently distorts nearby content, while ObjectClear may over-smooth or alter surrounding structures, as seen in the first row of Fig.~\ref{fig:qual_ober} and the third column of Fig.~\ref{fig:qual_rord}. In contrast, our method more consistently confines edits to the masked region, reconstructs semantically plausible content, and preserves the unmasked scene.

These results highlight the benefit of incorporating semantic prediction into inpainting. Rather than relying only on low-level texture extrapolation, the predicted semantic prior for the hidden region guides the generator toward content consistent with the visible context. This is especially evident in challenging occlusion cases, such as the bicycle example of Fig.~\ref{fig:qual_rord}, where removing the person requires reconstructing occluded bicycle parts. Compared with competing methods, our approach recovers these structures more coherently.

Similarly, in the OBER-Test examples of Fig.~\ref{fig:qual_ober}, which contain object-induced effects such as reflections and shadows, our model removes both the target object and its associated effect when the corresponding mask is provided, while preserving coherent background content. Overall, these qualitative results indicate that the proposed semantic-guided framework enables context-aware object-and-effect removal with minimal disturbance to unmasked regions.

\subsubsection{Efficiency and Run-time Performance} 
\label{subsubsec:time_size}

\begin{figure}[htbp]
  \centering
  \includegraphics[width=0.95\linewidth]{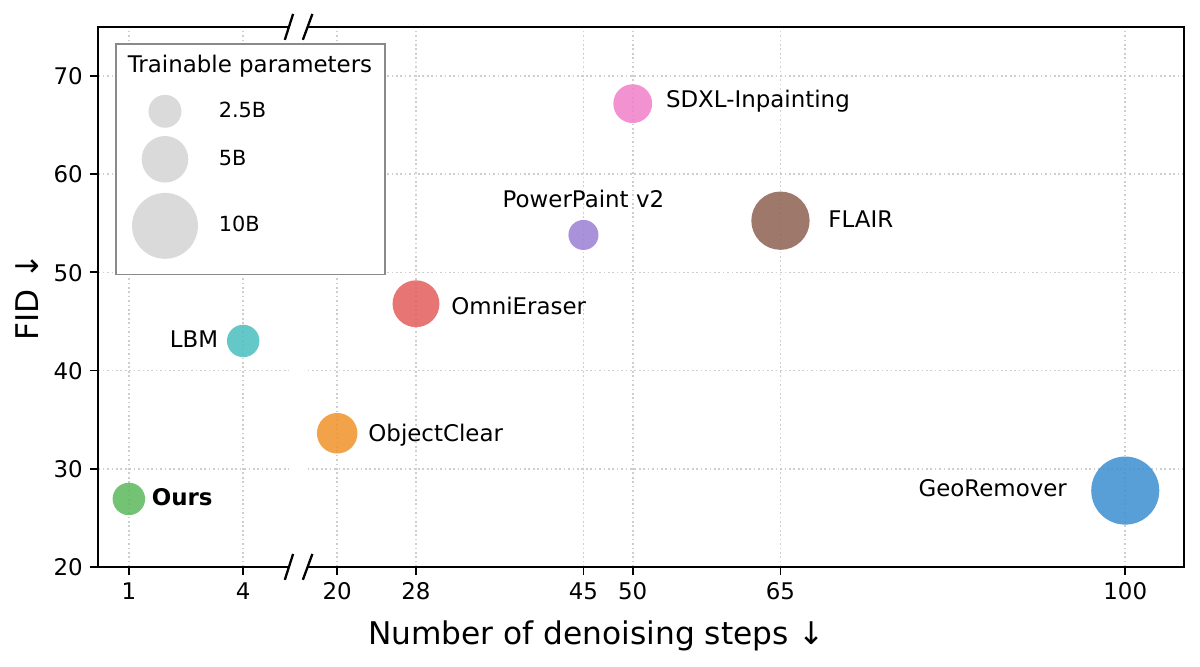}
  \caption{Quality--efficiency comparison on the RORD-Val dataset. FID is plotted against the number of denoising steps; lower is better for both axes. SERO achieves the best trade-off, attaining the lowest FID with only a single denoising step.}
  \Description{Scatter plot of FID versus denoising steps. SERO appears at one step with the lowest FID, while competing methods use between 20 and 100 steps.}
  \label{fig:size_comparisons}
\end{figure}

Fig.~\ref{fig:size_comparisons} and Table~\ref{tab:time_size_comparison} compare image quality, inference efficiency, runtime, and trainable model size on RORD-Val and a shared 50-sample benchmark. SERO achieves the best FID of 26.92 with a single denoising step, whereas competing methods require 20--100 steps; GeoRemover, for example, reaches the closest FID of 27.77 using 100 steps. All runtime measurements were performed on a single NVIDIA H200 GPU to ensure a common evaluation setting while accommodating the high memory requirements of several competing methods. Under this setup, SERO runs in $0.321\,\mathrm{s}$ per sample, making it $4.5\times$ faster than ObjectClear and $83.1\times$ faster than GeoRemover, while retaining a compact trainable parameter count. Compared with our LBM backbone, which uses the same architecture and $2.43$B trainable parameters but requires four bridge steps and is $1.7\times$ slower, SERO shows that the semantic prior eliminates iterative refinement without additional runtime overhead, since the frozen prior requires only a single forward pass. This combination of quality and efficiency makes SERO well suited for resource-constrained deployment, including mobile and edge devices. 

\subsubsection{User Study} 
\label{subsubsec:user_study}
Object removal is inherently ill-posed, as multiple visually plausible and semantically consistent completions may exist for the same masked region. Consequently, direct comparison with a single ground-truth image and conventional pixel- or distribution-based metrics may not fully reflect the perceptual quality of the generated result. To complement our quantitative evaluation and obtain a more direct assessment of visual realism and removal quality, we therefore conduct a user study in which participants compare the outputs of \textsc{SERO} with those of competing methods.

\begin{table}[htbp]
\centering
\scriptsize
\setlength{\tabcolsep}{5pt}
\caption{\textbf{Inference efficiency and model size.} SERO achieves the fastest runtime with single-step inference.}
\renewcommand{\arraystretch}{1.0}
\begin{tabular}{l c r r r}
\toprule
\textbf{Method} & \textbf{Steps} & \textbf{Median (s)} & \textbf{Slowdown} & \textbf{Trainable (B)} \\
\midrule
LBM             & 4            & 0.530  & $1.7\times$  & 2.43  \\
ObjectClear             & 20           & 1.442  & $4.5\times$  & 3.79  \\
SDXL-Inpainting         & 50           & 2.342  & $7.3\times$  & 3.47  \\
PowerPaint~v2           & 45           & 6.549  & $20.4\times$ & \textbf{2.08}  \\
OmniEraser              & 28           & 6.832  & $21.3\times$ & 5.06  \\
FLAIR                   & 65           & 11.464 & $35.7\times$ & 7.83  \\
GeoRemover & $50{+}50$ & 26.685 & $83.1\times$ & 10.68 \\
\midrule
\textbf{Ours}            & \textbf{1}   & \textbf{0.321} & \textbf{1.0$\times$} & 2.43 \\
\bottomrule
\end{tabular}

\label{tab:time_size_comparison}
\end{table}

\begin{figure}[t]  
  \centering
  \includegraphics[width=0.98\linewidth]{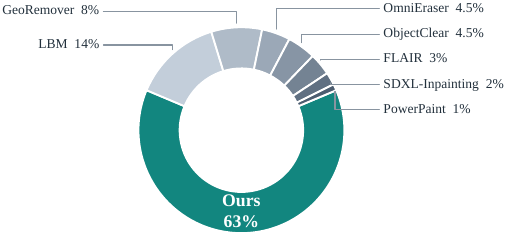}
  \Description{A pie chart showing that our method, SERO, is highly preferred by users compared to existing object-removal baselines.}
  \caption{User study preference distribution across object-removal methods.}
  \label{fig:user}
\end{figure}

In a blind user study, 20 participants evaluated 10 images by comparing object-removal results while viewing the input image with the corresponding mask overlaid. As shown in \cref{fig:user}, our method achieved the highest perceptual preference, accounting for 63\% of the recorded selections, substantially outperforming LBM (14\%), GeoRemover (8\%), OmniEraser (4.5\%), ObjectClear (4.5\%), FLAIR (3.5\%), SDXL-Inpainting (2\%), and PowerPaint (1\%). The study suggests that our method produces outputs that are more frequently judged as visually plausible by human observers.

\subsection{Ablation Analysis}
\label{sec:ablation}

In this section, we analyse the contribution of key components of our method. We provide ablations on the predictive semantic prior and the number of sampling steps in the main paper, while additional ablations on the loss function and architectural design are included in the supplementary material. 

\subsubsection{Semantic prior conditioning}
\begin{figure}[!ht]
  \centering
  \includegraphics[width=0.95\linewidth]{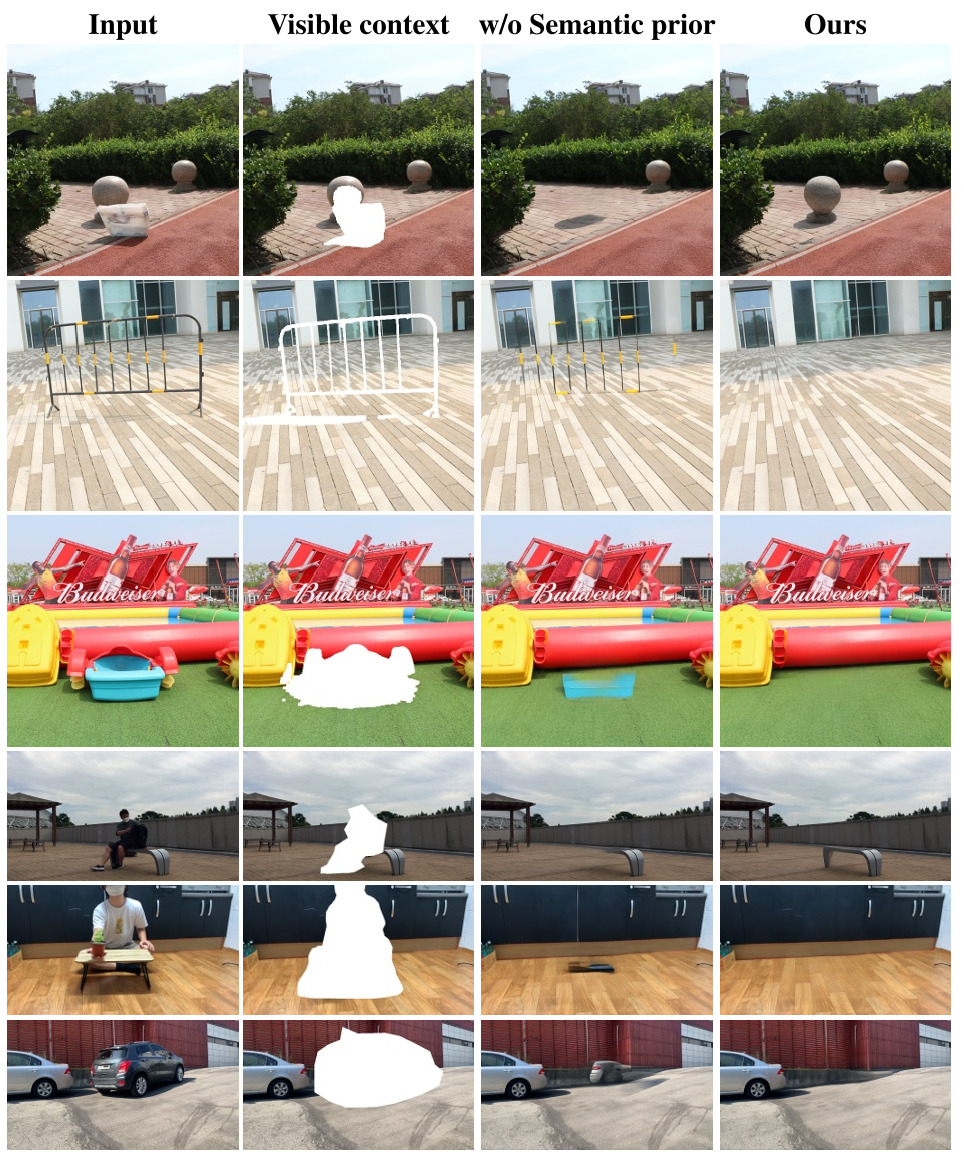}
  \caption{\textbf{Qualitative effect of the semantic prior.} Compared with the baseline, SERO produces more contextually coherent completions by conditioning bridge matching on the predicted hidden-region structure.}
  \Description{Object-removal examples comparing the input, visible context, a baseline without a semantic prior, and SERO with the semantic prior.}
  \label{fig:abl_qualitative}
\end{figure}

Table~\ref{tab:abl_conditioning} studies whether the gains from semantic conditioning arise specifically from I-JEPA's predictive objective or simply from concatenating pretrained visual features to the model input. All variants use the same spatial conditioning and feature-concatenation strategy, differing only in the source of the semantic features. Without a semantic prior, \textsc{SERO} obtains an FID of 33.03 and local FID of 43.23. Conditioning on features from CLIP~\cite{radford2021learning} or DINOv2~\cite{oquab2023dinov2} improves performance, confirming that additional semantic representations are beneficial. However, I-JEPA consistently performs best across all metrics, achieving an FID of 28.02 and fMSE of 807.38, compared with FIDs of 29.45 and 29.25 and fMSEs of 867.64 and 878.38 for CLIP and DINOv2, respectively. This suggests that the improvement is not solely due to providing richer visual features. Unlike CLIP and DINOv2, which primarily encode visible image content, I-JEPA explicitly learns to predict the semantic representation of a hidden region from its surrounding context, closely matching the object-removal setting in which the model must infer what should plausibly occupy the masked area. The qualitative comparisons in Fig.~\ref{fig:abl_qualitative} reinforce the quantitative results, showing that I-JEPA conditioning produces fewer removal artifacts and more semantically coherent completions than the alternative feature extractors.

\begin{table}[h]
\centering
\scriptsize
\setlength{\tabcolsep}{4.0pt}
\renewcommand{\arraystretch}{1.0}
\caption{\textbf{Semantic prior ablation.} I-JEPA outperforms CLIP and DINOv2 conditioning across all metrics.}
\resizebox{\linewidth}{!}{%
\begin{tabular}{lccccc}
\toprule
\textbf{Semantic Conditioning} & \textbf{FID}$\downarrow$ & \textbf{localFID}$\downarrow$ & \textbf{fMSE}$\downarrow$ & \textbf{PSNR}$\uparrow$ & \textbf{SSIM (\%)}$\uparrow$ \\
\midrule
Base (no semantic prior) & 33.03 & 43.23 & 966.39 & 21.47 & 67.32 \\
\midrule
Base + CLIP features & 29.45 & 40.19 & \underline{867.64} & \underline{21.85} & \underline{67.49} \\
Base + DINOv2 features & \underline{29.25} & \underline{40.14} & 878.38 & 21.84 & 67.42 \\
Base + I-JEPA predictive prior \textbf{(Ours)} & \textbf{28.02} & \textbf{37.95} & \textbf{807.38} & \textbf{22.13} & \textbf{67.66} \\
\bottomrule
\end{tabular}%
}

\label{tab:abl_conditioning}
\end{table}

\begin{table}[th]
\centering
{\fontsize{6.9}{7.5}\selectfont
\setlength{\tabcolsep}{4pt}
\renewcommand{\arraystretch}{1.0}
\caption{\textbf{Effect of inference steps.} One-step inference provides the best accuracy--efficiency trade-off, with additional steps yielding no consistent gains. \label{tab:abl_steps}}
\begin{tabular}{cccccc}
\toprule
\textbf{Number of Steps} & \textbf{FID}$\downarrow$ & \textbf{localFID}$\downarrow$ & \textbf{fMSE}$\downarrow$ & \textbf{PSNR}$\uparrow$ & \textbf{SSIM(\%)}$\uparrow$ \\
\midrule
\textbf{1 (ours)} & \textbf{28.02} & 37.95 & \textbf{807.38} & \textbf{22.13} & \textbf{67.66} \\
2 & 28.67 & \textbf{37.92} & 853.08 & 21.94 & 67.25 \\
3 & 40.34 & 58.28 & 929.65 & 21.92 & 67.11 \\
4 & 31.25 & 40.63 & 924.70 & 21.69 & 66.78 \\
\bottomrule
\end{tabular}%
}
\end{table}

\subsubsection{Number of inference steps}
We evaluate SERO with 1 through 4 inference steps on the validation set in Table~\ref{tab:abl_steps}. A single step gives the best overall performance, achieving the best FID, fMSE, PSNR, and SSIM, while two steps improve localFID by only a small fraction of $0.03$. Additional steps do not improve quality and often degrade it. 

SERO achieves its best accuracy--efficiency trade-off in one forward pass, suggesting that the semantic prior reduces the need for iterative refinement. By contrast, DDIM-based diffusion \cite{zhuang2024powerpaint, podell2024sdxl, zhao2026objectclear, zhu2025georemover} models typically require at least 20 sampling steps to reach their best quality.

\section{Limitations}

While our model achieves strong object-removal performance, its reliability depends on the quality of the input mask and the segmentation model used to obtain it. Inaccurate or coarse masks can remove useful context or leave foreground remnants, making realistic completion harder. Performance may also degrade when the removed object covers a large region, requiring the model to hallucinate uncertain content from limited evidence. Finally, occluded text remains challenging because it requires exact reconstruction; plausible completions may still fail to preserve the correct characters or wording.

\section{Conclusion}

We introduced \textsc{SERO}, a predictive semantic-prior bridge matching framework for semantic object removal and context-aware image inpainting. By combining contextual semantic prediction with an efficient latent bridge-matching backbone, our model infers plausible missing content while removing target objects and their associated visual effects, including shadows, reflections, and occlusion artifacts. Guided by the predicted semantic prior, our method produces photorealistic and semantically coherent completions in as few as one generation step. Extensive experiments demonstrate strong quantitative and qualitative performance, highlighting SERO’s potential for practical, robust, and semantically aware object removal.

\begin{acks}
This work has been partially funded by the EU projects VICTOR-XR (GA: Nr 101298864) and SUBLIMITY (GA: Nr 101294363).
\end{acks}

\bibliographystyle{ACM-Reference-Format}
\bibliography{main}

@String{Computer = "{IEEE} Computer" }

@inproceedings{assran2023self,
  title={Self-supervised learning from images with a joint-embedding predictive architecture},
  author={Assran, Mahmoud and Duval, Quentin and Misra, Ishan and Bojanowski, Piotr and Vincent, Pascal and Rabbat, Michael and LeCun, Yann and Ballas, Nicolas},
  booktitle={2023 IEEE/CVF Conference on Computer Vision and Pattern Recognition (CVPR)},
  pages={15619--15629},
  year={2023},
  organization={IEEE}
}

@inproceedings{zhang2018unreasonable,
  title={The unreasonable effectiveness of deep features as a perceptual metric},
  author={Zhang, Richard and Isola, Phillip and Efros, Alexei A and Shechtman, Eli and Wang, Oliver},
  booktitle={2018 IEEE/CVF conference on computer vision and pattern recognition},
  pages={586--595},
  year={2018},
  organization={IEEE}
}

@inproceedings{xie2023smartbrush,
  title     = {SmartBrush: Text and Shape Guided Object Inpainting with Diffusion Model},
  author    = {Xie, Shaoan and Zhang, Zhifei and Lin, Zhe and Hinz, Tobias and Zhang, Kun},
  booktitle = {Proceedings of the IEEE/CVF Conference on Computer Vision and Pattern Recognition},
  pages     = {22428--22437},
  year      = {2023}
}

@inproceedings{hore2010image,
  title     = {Image Quality Metrics: {PSNR} vs. {SSIM}},
  author    = {Hor{\'e}, Alain and Ziou, Djemel},
  booktitle = {2010 20th International Conference on Pattern Recognition},
  pages     = {2366--2369},
  year      = {2010},
  organization = {IEEE}
}

@inproceedings{podell2024sdxl,
  title     = {{SDXL}: Improving Latent Diffusion Models for High-Resolution Image Synthesis},
  author    = {Podell, Dustin and English, Zion and Lacey, Kyle and Blattmann, Andreas and Dockhorn, Tim and M{\"u}ller, Jonas and Penna, Joe and Rombach, Robin},
  booktitle = {The Twelfth International Conference on Learning Representations},
  year      = {2024}
}

@article{oquab2023dinov2,
  title={DINOv2: Learning Robust Visual Features without Supervision},
  author={Oquab, Maxime and Darcet, Timothee and Mrazek, Theo and others},
  journal={arXiv preprint arXiv:2304.07193},
  year={2023}
}

@article{ho2020denoising,
  title={Denoising diffusion probabilistic models},
  author={Ho, Jonathan and Jain, Ajay and Abbeel, Pieter},
  journal={Advances in Neural Information Processing Systems},
  volume={33},
  pages={6840--6851},
  year={2020}
}

@article{goodfellow2014generative,
  title={Generative adversarial nets},
  author={Goodfellow, Ian and Pouget-Abadie, Jean and Mirza, Mehdi and Xu, Bing and Warde-Farley, David and Ozair, Sherjil and Courville, Aaron and Bengio, Yoshua},
  journal={Advances in Neural Information Processing Systems},
  volume={27},
  year={2014}
}

@inproceedings{rombach2022high,
  title={High-resolution image synthesis with latent diffusion models},
  author={Rombach, Robin and Blattmann, Andreas and Lorenz, Dominik and Esser, Patrick and Ommer, Bj{\"o}rn},
  booktitle={Proceedings of the IEEE/CVF conference on computer vision and pattern recognition},
  pages={10684--10695},
  year={2022}
}

@inproceedings{radford2021learning,
  title={Learning transferable visual models from natural language supervision},
  author={Radford, Alec and Kim, Jong Wook and Hallacy, Chris and Ramesh, Aditya and Goh, Gabriel and Agarwal, Sandhini and Sastry, Girish and Askell, Amanda and Mishkin, Pamela and Clark, Jack and others},
  booktitle={International conference on machine learning},
  pages={8748--8763},
  year={2021},
  organization={PMLR}
}

@inproceedings{esser2024scaling,
  title={Scaling rectified flow transformers for high-resolution image synthesis},
  author={Esser, Patrick and Kulal, Sumith and Blattmann, Andreas and Entezari, Rahim and M{\"u}ller, Jonas and Saini, Harry and Levi, Yam and Lorenz, Dominik and Sauer, Axel and Boesel, Frederic and others},
  booktitle={Forty-first international conference on machine learning},
  year={2024}
}

@article{labs2025flux,
  title={FLUX. 1 Kontext: Flow Matching for In-Context Image Generation and Editing in Latent Space},
  author={Labs, Black Forest and Batifol, Stephen and Blattmann, Andreas and Boesel, Frederic and Consul, Saksham and Diagne, Cyril and Dockhorn, Tim and English, Jack and English, Zion and Esser, Patrick and others},
  journal={arXiv preprint arXiv:2506.15742},
  year={2025}
}

@article{lipman2022flow,
  title={Flow matching for generative modeling},
  author={Lipman, Yaron and Chen, Ricky TQ and Ben-Hamu, Heli and Nickel, Maximilian and Le, Matt},
  journal={arXiv preprint arXiv:2210.02747},
  year={2022}
}

@inproceedings{chadebec2025lbm,
  title={Lbm: Latent bridge matching for fast image-to-image translation},
  author={Chadebec, Cl{\'e}ment and Tasar, Onur and Sreetharan, Sanjeev and Aubin, Benjamin},
  booktitle={Proceedings of the IEEE/CVF International Conference on Computer Vision},
  pages={29086--29098},
  year={2025}
}

@inproceedings{heusel2017gans,
  title={GANs trained by a two time-scale update rule converge to a local Nash equilibrium},
  author={Heusel, Martin and Ramsauer, Hubert and Unterthiner, Thomas and Nessler, Bernhard and Hochreiter, Sepp},
  booktitle={Advances in Neural Information Processing Systems},
  year={2017}
}

@inproceedings{pathak2016context,
  title     = {Context Encoders: Feature Learning by Inpainting},
  author    = {Pathak, Deepak and Krahenbuhl, Philipp and Donahue, Jeff and Darrell, Trevor and Efros, Alexei A.},
  booktitle = {Proceedings of the IEEE Conference on Computer Vision and Pattern Recognition},
  year      = {2016}
}

@inproceedings{liu2020rethinking,
  title     = {Rethinking Image Inpainting via a Mutual Encoder-Decoder with Feature Equalizations},
  author    = {Liu, Hongyu and Jiang, Bin and Song, Yibing and Huang, Wei and Yang, Chao},
  booktitle = {Proceedings of the European Conference on Computer Vision},
  year      = {2020}
}

@inproceedings{ren2019structureflow,
  title     = {StructureFlow: Image Inpainting via Structure-Aware Appearance Flow},
  author    = {Ren, Yurui and Yu, Xiaoming and Zhang, Ruonan and Li, Thomas H. and Liu, Shan and Li, Ge},
  booktitle = {Proceedings of the IEEE International Conference on Computer Vision},
  year      = {2019}
}

@inproceedings{zeng2019learning,
  title     = {Learning Pyramid-Context Encoder Network for High-Quality Image Inpainting},
  author    = {Zeng, Yanhong and Fu, Jianlong and Chao, Hongyang and Guo, Baining},
  booktitle = {Proceedings of the IEEE Conference on Computer Vision and Pattern Recognition},
  year      = {2019}
}

@article{song2020score,
  title   = {Score-Based Generative Modeling through Stochastic Differential Equations},
  author  = {Song, Yang and Sohl-Dickstein, Jascha and Kingma, Diederik P. and Kumar, Abhishek and Ermon, Stefano and Poole, Ben},
  journal = {arXiv preprint arXiv:2011.13456},
  year    = {2020}
}

@article{meng2021sdedit,
  title   = {{SDEdit}: Guided Image Synthesis and Editing with Stochastic Differential Equations},
  author  = {Meng, Chenlin and He, Yutong and Song, Yang and Song, Jiaming and Wu, Jiajun and Zhu, Jun-Yan and Ermon, Stefano},
  journal = {arXiv preprint arXiv:2108.01073},
  year    = {2021}
}

@inproceedings{lugmayr2022repaint,
  title     = {{RePaint}: Inpainting using Denoising Diffusion Probabilistic Models},
  author    = {Lugmayr, Andreas and Danelljan, Martin and Romero, Andres and Yu, Fisher and Timofte, Radu and Van Gool, Luc},
  booktitle = {Proceedings of the IEEE Conference on Computer Vision and Pattern Recognition},
  year      = {2022}
}

@inproceedings{avrahami2022blended,
  title     = {Blended Diffusion for Text-Driven Editing of Natural Images},
  author    = {Avrahami, Omri and Lischinski, Dani and Fried, Ohad},
  booktitle = {Proceedings of the IEEE Conference on Computer Vision and Pattern Recognition},
  year      = {2022}
}

@inproceedings{zhuang2024powerpaint,
  title     = {A Task is Worth One Word: Learning with Task Prompts for High-Quality Versatile Image Inpainting},
  author    = {Zhuang, Junhao and Zeng, Yanhong and Liu, Wenran and Yuan, Chun and Chen, Kai},
  booktitle = {Proceedings of the European Conference on Computer Vision},
  year      = {2024}
}

@inproceedings{ju2024brushnet,
  title     = {{BrushNet}: A Plug-and-Play Image Inpainting Model with Decomposed Dual-Branch Diffusion},
  author    = {Ju, Xuan and Liu, Xian and Wang, Xintao and Bian, Yuxuan and Shan, Ying and Xu, Qiang},
  booktitle = {Proceedings of the European Conference on Computer Vision},
  year      = {2024}
}

@inproceedings{wang2025enhanced,
  title     = {Towards Enhanced Image Inpainting: Mitigating Unwanted Object Insertion and Preserving Color Consistency},
  author    = {Wang, Yikai and Cao, Chenjie and Yu, Junqiu and Fan, Ke and Xue, Xiangyang and Fu, Yanwei},
  booktitle = {Proceedings of the IEEE Conference on Computer Vision and Pattern Recognition},
  year      = {2025}
}

@inproceedings{chen2024freecompose,
  title     = {{FreeCompose}: Generic Zero-Shot Image Composition with Diffusion Prior},
  author    = {Chen, Zhekai and Wang, Wen and Yang, Zhen and Yuan, Zeqing and Chen, Hao and Shen, Chunhua},
  booktitle = {Proceedings of the European Conference on Computer Vision},
  year      = {2024}
}

@inproceedings{jia2025designedit,
  title     = {{DesignEdit}: Multi-Layered Latent Decomposition and Fusion for Unified and Accurate Image Editing},
  author    = {Jia, Yueru and Yuan, Yuhui and Cheng, Aosong and Wang, Chuke and Li, Ji and Jia, Huizhu and Zhang, Shanghang},
  booktitle = {Proceedings of the AAAI Conference on Artificial Intelligence},
  year      = {2025}
}

@inproceedings{sun2025attentive,
  title     = {Attentive Eraser: Unleashing Diffusion Model's Object Removal Potential via Self-Attention Redirection Guidance},
  author    = {Sun, Wenhao and Cui, Benlei and Dong, Xue-Mei and Tang, Jingqun},
  booktitle = {Proceedings of the AAAI Conference on Artificial Intelligence},
  year      = {2025}
}

@inproceedings{ekin2024clipaway,
  title     = {{CLIPAway}: Harmonizing Focused Embeddings for Removing Objects via Diffusion Models},
  author    = {Ekin, Yigit and Yildirim, Ahmet Burak and Caglar, Erdem Eren and Erdem, Aykut and Erdem, Erkut and Dundar, Aysegul},
  booktitle = {Advances in Neural Information Processing Systems},
  year      = {2024}
}

@inproceedings{jiang2025smarteraser,
  title     = {{SmartEraser}: Remove Anything from Images using Masked-Region Guidance},
  author    = {Jiang, Longtao and Wang, Zhendong and Bao, Jianmin and Zhou, Wengang and Chen, Dongdong and Shi, Lei and Chen, Dong and Li, Houqiang},
  booktitle = {Proceedings of the IEEE Conference on Computer Vision and Pattern Recognition},
  year      = {2025}
}

@inproceedings{liu2025erasediffusion,
  title     = {Erase Diffusion: Empowering Object Removal through Calibrating Diffusion Pathways},
  author    = {Liu, Yi and Zhou, Hao and Shang, Wenxiang and Lin, Ran and Cui, Benlei},
  booktitle = {Proceedings of the IEEE Conference on Computer Vision and Pattern Recognition},
  year      = {2025}
}

@inproceedings{li2025rorem,
  title     = {{RORem}: Training a Robust Object Remover with Human-in-the-Loop},
  author    = {Li, Ruibin and Yang, Tao and Guo, Song and Zhang, Lei},
  booktitle = {Proceedings of the IEEE Conference on Computer Vision and Pattern Recognition},
  year      = {2025}
}

@inproceedings{yu2025omnipaint,
  title     = {{OmniPaint}: Mastering Object-Oriented Editing via Disentangled Insertion-Removal Inpainting},
  author    = {Yu, Yongsheng and Zeng, Ziyun and Zheng, Haitian and Luo, Jiebo},
  booktitle = {Proceedings of the IEEE International Conference on Computer Vision},
  year      = {2025}
}

@article{wei2025omnieraser,
  title   = {{OmniEraser}: Remove Objects and Their Effects in Images with Paired Video-Frame Data},
  author  = {Wei, Runpu and Yin, Zijin and Zhang, Shuo and Zhou, Lanxiang and Wang, Xueyi and Ban, Chao and Cao, Tianwei and Sun, Hao and He, Zhongjiang and Liang, Kongming and Ma, Zhanyu},
  journal = {arXiv preprint arXiv:2501.07397},
  year    = {2025}
}

@inproceedings{zhu2025georemover,
  title     = {{GeoRemover}: Removing Objects and Their Causal Visual Artifacts},
  author    = {Zhu, Zixin and Li, Haoxiang and Feng, Xuelu and Wu, He and Qiao, Chunming and Yuan, Junsong},
  booktitle = {Advances in Neural Information Processing Systems},
  year      = {2025}
}

@inproceedings{erbach2026solving,
  title     = {Solving Inverse Problems with {FLAIR}},
  author    = {Erbach, Julius and Narnhofer, Dominik and Dombos, Andreas and Schiele, Bernt and Lenssen, Jan Eric and Schindler, Konrad},
  booktitle = {Advances in Neural Information Processing Systems},
  volume    = {38},
  pages     = {136667--136704},
  year      = {2025}
}

@InProceedings{zhao2026objectclear,
      title     = {Precise Object and Effect Removal with Adaptive Target-Aware Attention},
      author    = {Zhao, Jixin and Wang, Zhouxia and Yang, Peiqing and Zhou, Shangchen},
      booktitle = {CVPR},
      year      = {2026},
    }

@inproceedings{Sagong_2022_BMVC,
  author    = {Min-Cheol Sagong and Yoon-Jae Yeo and Seung-Won Jung and Sung-Jea Ko},
  title     = {RORD: A Real-world Object Removal Dataset},
  booktitle = {Proceedings of the British Machine Vision Conference (BMVC)},
  year      = {2022}
}

@inproceedings{peebles2023scalable,
  title={Scalable diffusion models with transformers},
  author={Peebles, William and Xie, Saining},
  booktitle={Proceedings of the IEEE/CVF international conference on computer vision},
  pages={4195--4205},
  year={2023}
}

@inproceedings{liu2025effective,
  title={Effective cloud removal for remote sensing images by an improved mean-reverting denoising model with elucidated design space},
  author={Liu, Yi and Li, Wengen and Guan, Jihong and Zhou, Shuigeng and Zhang, Yichao},
  booktitle={2025 IEEE/CVF Conference on Computer Vision and Pattern Recognition (CVPR)},
  pages={17851--17861},
  year={2025},
  organization={IEEE}
}

@article{wang2024idf,
  title={IDF-CR: Iterative diffusion process for divide-and-conquer cloud removal in remote-sensing images},
  author={Wang, Meilin and Song, Yexing and Wei, Pengxu and Xian, Xiaoyu and Shi, Yukai and Lin, Liang},
  journal={IEEE Transactions on Geoscience and Remote Sensing},
  volume={62},
  pages={1--14},
  year={2024},
  publisher={IEEE}
}

@inproceedings{yu2018generative,
  title={Generative image inpainting with contextual attention},
  author={Yu, Jiahui and Lin, Zhe and Yang, Jimei and Shen, Xiaohui and Lu, Xin and Huang, Thomas S},
  booktitle={Proceedings of the IEEE conference on computer vision and pattern recognition},
  pages={5505--5514},
  year={2018}
}

@InProceedings{Nazeri_2019_ICCV,
author = {Nazeri, Kamyar and Ng, Eric and Joseph, Tony and Qureshi, Faisal and Ebrahimi, Mehran},
title = {EdgeConnect: Structure Guided Image Inpainting using Edge Prediction},
booktitle = {Proceedings of the IEEE/CVF International Conference on Computer Vision (ICCV) Workshops},
month = {Oct},
year = {2019}
}

@inproceedings{suvorov2022resolution,
  title={Resolution-robust large mask inpainting with fourier convolutions},
  author={Suvorov, Roman and Logacheva, Elizaveta and Mashikhin, Anton and Remizova, Anastasia and Ashukha, Arsenii and Silvestrov, Aleksei and Kong, Naejin and Goka, Harshith and Park, Kiwoong and Lempitsky, Victor},
  booktitle={Proceedings of the IEEE/CVF winter conference on applications of computer vision},
  pages={2149--2159},
  year={2022}
}

@inproceedings{wang2018high,
  title={High-Resolution Image Synthesis and Semantic Manipulation with Conditional GANs},
  author={Wang, Ting-Chun and Liu, Ming-Yu and Zhu, Jun-Yan and Tao, Andrew and Kautz, Jan and Catanzaro, Bryan},
  booktitle={Proceedings of the IEEE Conference on Computer Vision and Pattern Recognition},
  year={2018}
}

@article{ren2024grounded,
  title={Grounded sam: Assembling open-world models for diverse visual tasks},
  author={Ren, Tianhe and Liu, Shilong and Zeng, Ailing and Lin, Jing and Li, Kunchang and Cao, He and Chen, Jiayu and Huang, Xinyu and Chen, Yukang and Yan, Feng and others},
  journal={arXiv preprint arXiv:2401.14159},
  year={2024}
}

\end{document}